\RequirePackage{xcolor}
\documentclass{bmvc2k}

\usepackage{booktabs} % For formal tables
\usepackage{subfig}

\newcommand{\mFIG}[1]{Fig.~(#1)}
\usepackage{xcolor,graphicx}
\usepackage[normalem]{ulem}

\usepackage{array}
\usepackage{arydshln}
\usepackage{graphicx,multicol}

\newcommand{\ignore}[1]{}
\newcommand{\bP}{{\mathbf {P}}}
\newcommand{\bh}{{\mathbf {h}}}

\title{Automatic Image Transformation for Inducing Affect}

\addauthor{Afsheen Rafaqat Ali}{afsheen.ali@itu.edu.pk}{1}
\addauthor{Mohsen Ali}{mohsen.ali@itu.edu.pk}{1}

\addinstitution{
 Intelligent Machines Lab\\
 Information Technology University of Punjab (ITU)\\
 Lahore, Pakistan
}

\runninghead{Ali et al.}{Automatic Image transformation for Inducing Affect}

\begin{document}

\maketitle

\begin{abstract}
Current image transformation and recoloring algorithms try to introduce artistic effects in the photographed images, based on user input of target image(s) or selection of pre-designed filters. These manipulations, although intended to enhance the impact of an image on the viewer, do not include the option of image transformation by specifying the affect information. In this paper we present an automatic image-transformation method that transforms the source image such that it can induce an emotional affect on the viewer, as desired by the user.
Our proposed novel image emotion transfer algorithm does not require a user-specified target image.
The proposed algorithm uses features  extracted from top layers of deep convolutional neural network and the user-specified emotion distribution to select multiple target images from an image database for color transformation, such that the resultant image has desired emotional impact.
Our method can handle more diverse set of photographs than the previous methods.
We conducted a detailed user study showing the effectiveness of our proposed method. A discussion and reasoning of failure cases has also been provided, indicating inherent limitation of color-transfer based methods in the use of emotion assignment.
\end{abstract}

\section{Introduction}
\label{sec:intro}
Over the last decade we have seen a dramatic increase in both the creation of multimedia content and its mass sharing. Social networks and content sharing websites make it easy to share photographs, consume the ones posted by others and to solicit the input from the viewers.
 Photographers use image enhancement techniques to either overcome the limitation of their sensors or to induce an artistic effect. Many existing applications allow easy manipulation of an image to garner more positive feedback from the viewers or enhance the context for sharing of personal experience.
Most of the currently available tools restrict themselves to a limited range of low-level filters (including smoothing and sharpening filters) and manipulations (e.g. giving vintage look to image, converting to black \& white or enhancing colors of landscape image). 
These manipulations are restricted to low level changes like adjusting hue, saturation and brightness, with user given minimum control over these parameters. 
Even the recently popular selfie or face filters introduced in commercial image sharing applications are limited to few designed alternations that could be applied to facial features.
%More recently advancement in deep learning has allowed tools that can transfer "style" from one image to other, allowing users to convert their photographs into paintings \citep{Gatys2016ImageST}. 

\begin{center}
\begin{figure}[t!]

\center
 
% -------------------------------------------------------------------------
 \begin{tabular}{m{2.0cm} m{2.0cm} m{2.0cm} m{2.0cm} m{2.0cm} m{2.0cm}} 
\small{Input Image} &   \hspace{-0.6cm} \small{Target Images} & \hspace{-0.8cm} \small{Output Image} & \hspace{-0.6cm} \small{Input Image} &   \hspace{-1.0cm} \small{Target Images} & \hspace{-1.4cm} \small{Output Image}\\
\end{tabular}
\hrule
 %---------------------------------------------------------------------------

\begin{tabular}{m{2.0cm} m{2.0cm} m{2.0cm} m{2.0cm} m{2.0cm} m{2.0cm}} 
  \vspace{-0.2cm}
  \subfloat{ \includegraphics[width=\textwidth, width=0.9\linewidth]{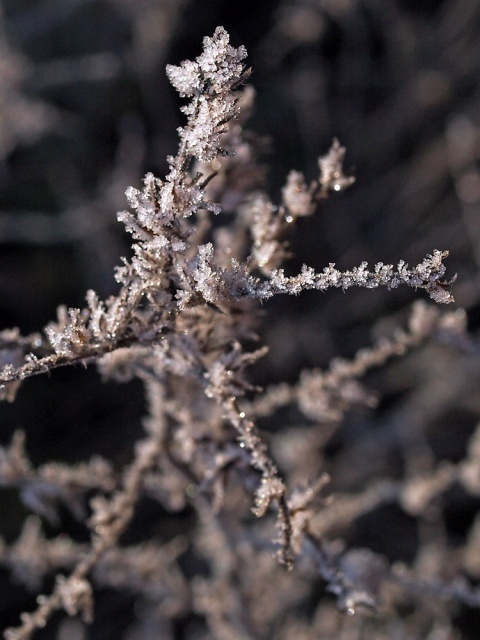}} &
   \vspace{-0.2cm}
  \hspace{-0.6cm}
 \subfloat{ \includegraphics[width=\textwidth, width=1\linewidth]{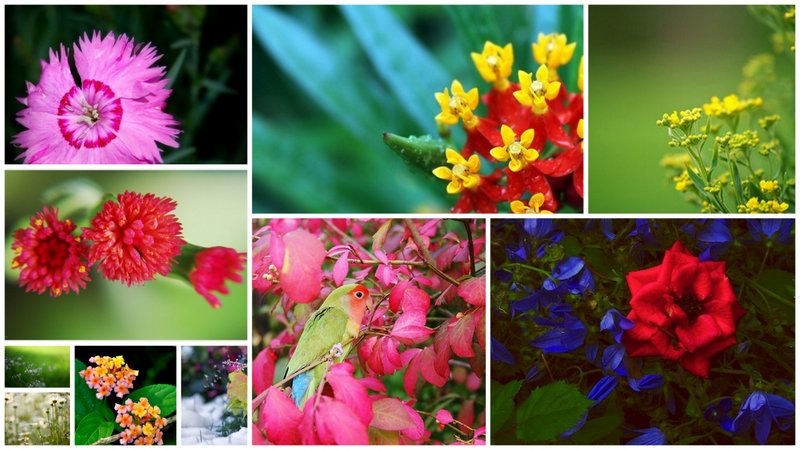}} &
   \vspace{-0.2cm}
  \hspace{-0.9cm}
 \subfloat{ \includegraphics[width=\textwidth, width=0.9\linewidth]{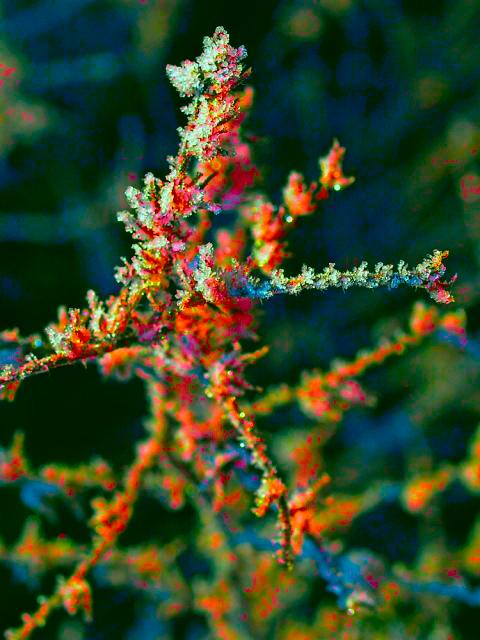}} &
   \vspace{-0.2cm}
  \hspace{-0.7cm}
  \subfloat{ \includegraphics[width=\textwidth, width=0.9\linewidth]{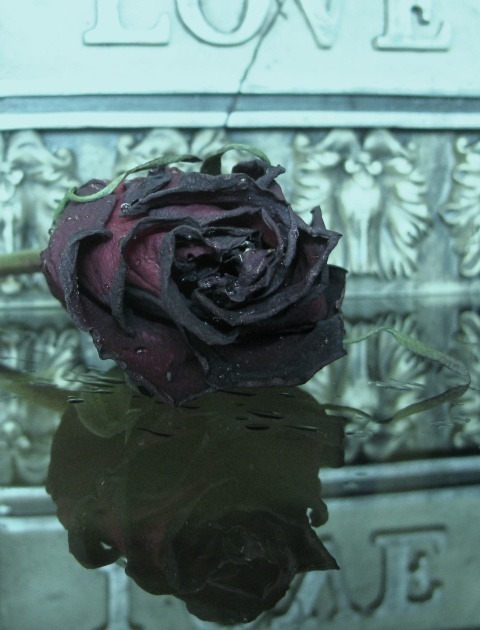}} &
   \vspace{-0.2cm}
  \hspace{-1.2cm}
 \subfloat{ \includegraphics[width=\textwidth, width=1\linewidth]{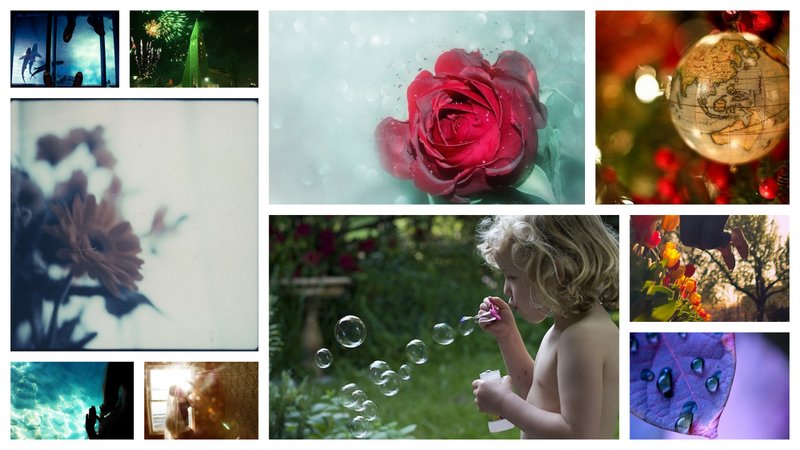}} &
  \vspace{-0.2cm}
  \hspace{-1.5cm}
 \subfloat{ \includegraphics[width=\textwidth, width=0.9\linewidth]{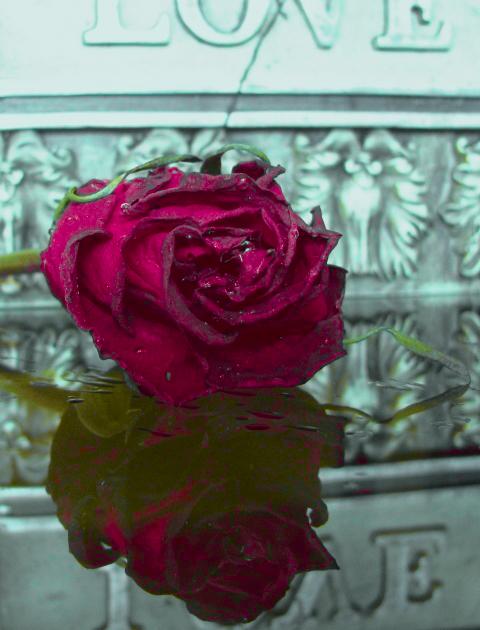}} 
 \\
 \vspace{-0.4cm} 
 \subfloat{\includegraphics[height=1.5cm, width=0.9\linewidth]{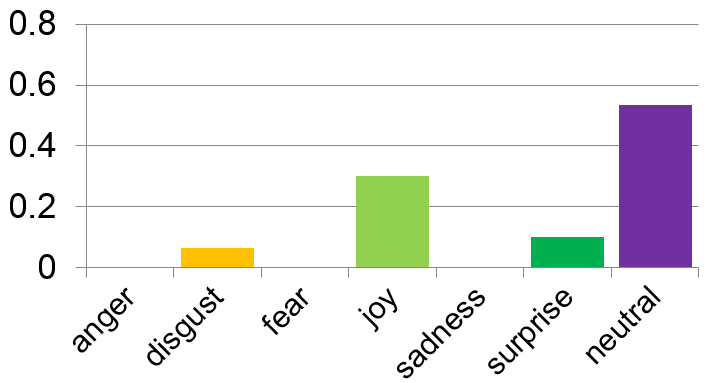}}  &
  \vspace{-0.4cm}
  \hspace{-0.6cm}
 \subfloat{\includegraphics[height=1.5cm, width=0.9\linewidth]{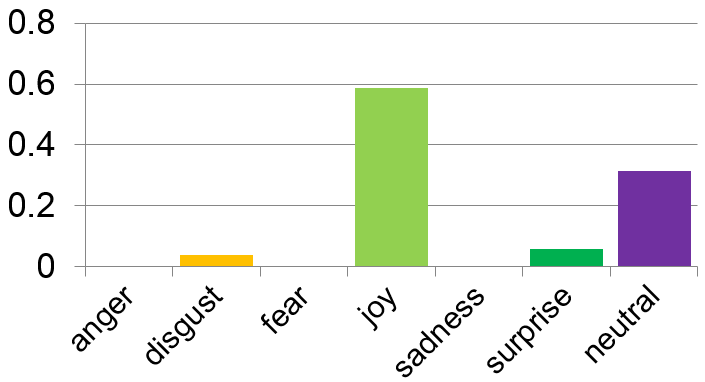}} & 
    \vspace{-0.6cm}
  \hspace{-0.9cm}
  Neutral $\rightarrow$ Joy &
 \vspace{-0.4cm}
  \hspace{-0.7cm} 
 \subfloat{\includegraphics[height=1.5cm, width=0.9\linewidth]{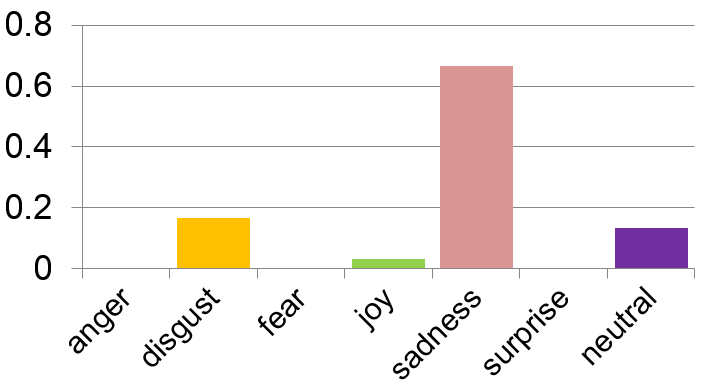}}  &
  \vspace{-0.4cm}
  \hspace{-1.1cm}
 \subfloat{\includegraphics[height=1.5cm, width=0.9\linewidth]{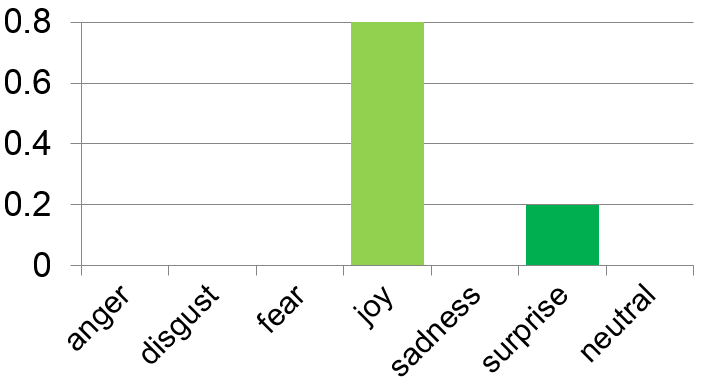}} & 
 \vspace{-0.8cm}
  \hspace{-1.2cm}
  Sad $\rightarrow$ Joy \vspace{-0.5cm}\\
  \cdashline{2-5}
 \end{tabular}
% -------------------------------------------------------------------------
 \begin{tabular}{m{2.0cm} m{2.0cm} m{2.0cm} m{2.0cm} m{2.0cm} m{2.0cm}} 
  \vspace{-0.2cm}
  \subfloat{ \includegraphics[width=\textwidth, width=0.9\linewidth]{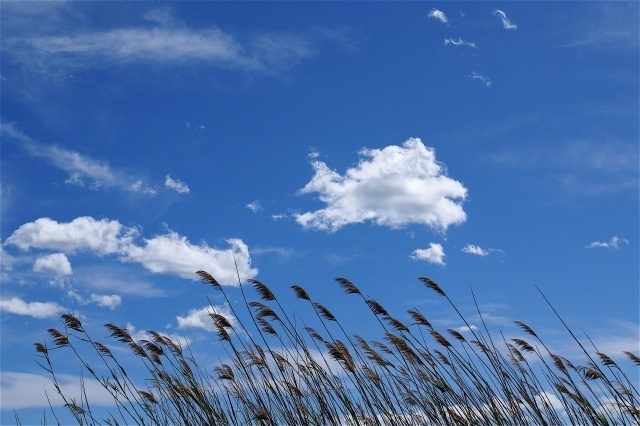}} &
   \vspace{-0.2cm}
  \hspace{-0.6cm}
 \subfloat{ \includegraphics[width=\textwidth, width=1\linewidth]{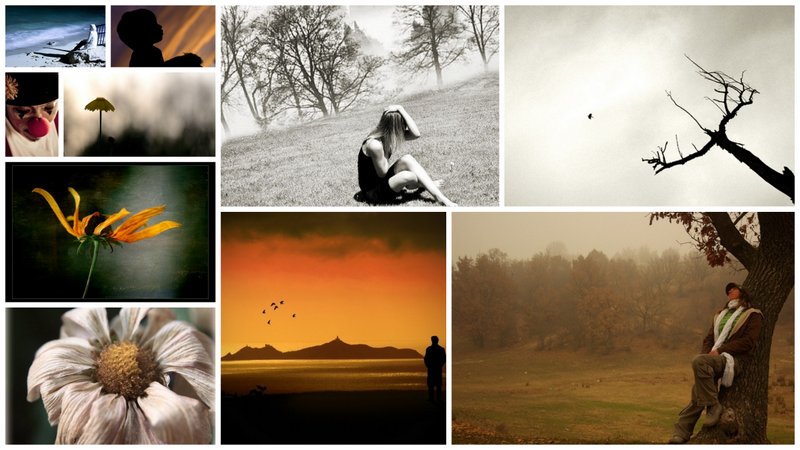}} &
   \vspace{-0.2cm}
  \hspace{-0.9cm}
 \subfloat{ \includegraphics[width=\textwidth, width=0.9\linewidth]{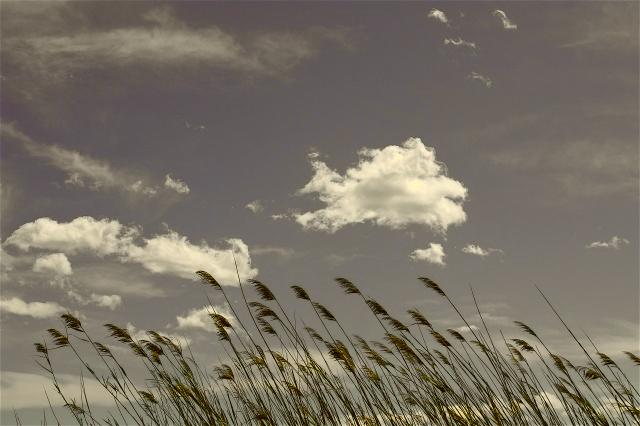}} &
   \vspace{-0.2cm}
  \hspace{-0.7cm}
  \subfloat{ \includegraphics[width=\textwidth, width=0.9\linewidth]{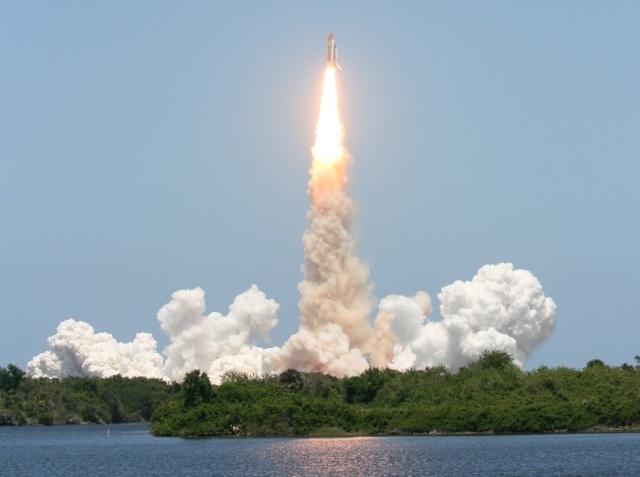}} &
   \vspace{-0.2cm}
  \hspace{-1.2cm}
 \subfloat{ \includegraphics[width=\textwidth, width=1\linewidth]{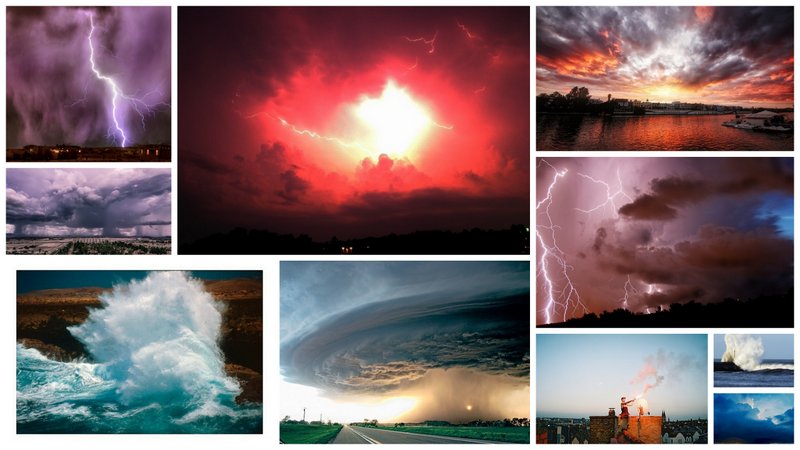}} &
  \vspace{-0.2cm}
  \hspace{-1.5cm}
 \subfloat{ \includegraphics[width=\textwidth, width=0.9\linewidth]{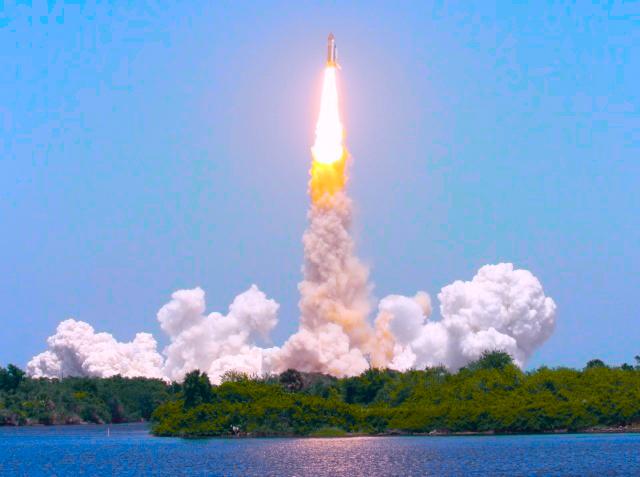}} 
 \\
 \vspace{-0.4cm} 
 \subfloat{\includegraphics[height=1.5cm, width=0.9\linewidth]{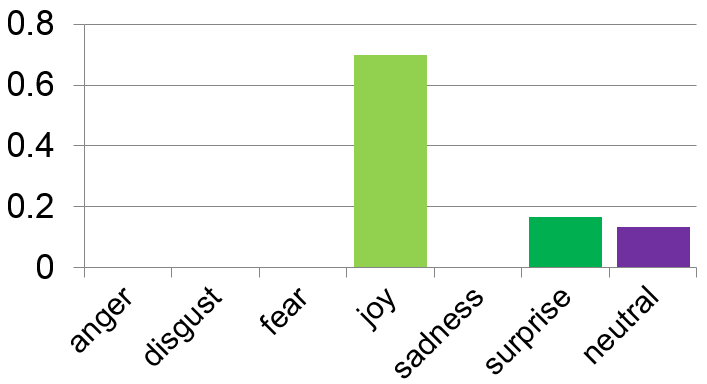}}  &
  \vspace{-0.4cm}
  \hspace{-0.6cm}
 \subfloat{\includegraphics[height=1.5cm, width=0.9\linewidth]{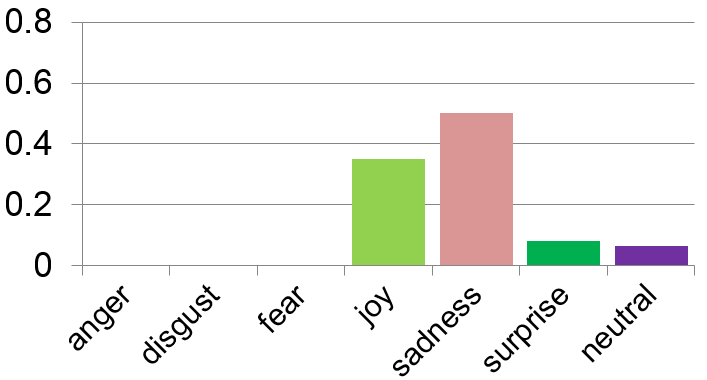}} & 
    \vspace{-0.6cm}
  \hspace{-0.9cm}
  Joy $\rightarrow$ Sad &
 \vspace{-0.4cm}
  \hspace{-0.7cm} 
 \subfloat{\includegraphics[height=1.5cm, width=0.9\linewidth]{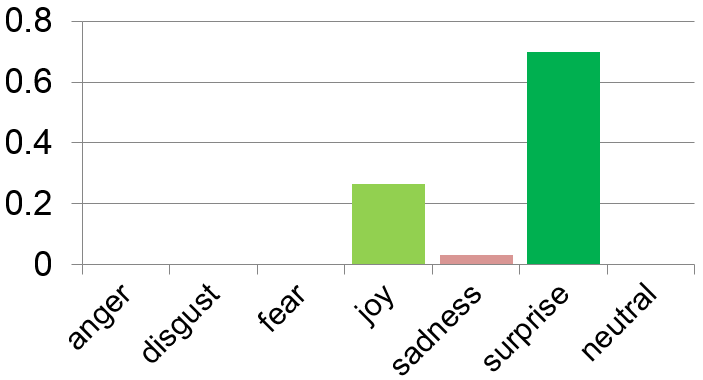}}  &
  \vspace{-0.4cm}
  \hspace{-1.1cm}
 \subfloat{\includegraphics[height=1.5cm, width=0.9\linewidth]{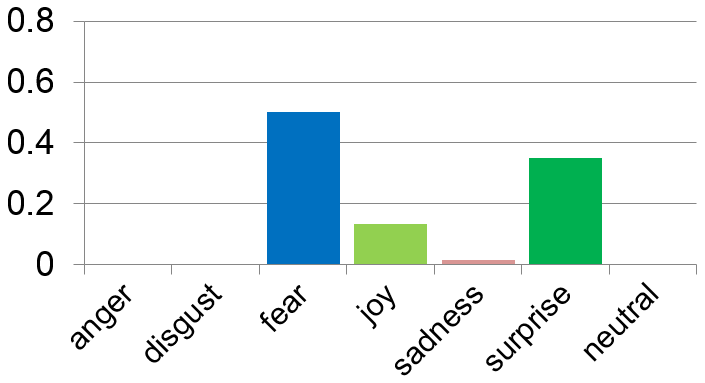}} & 
 \vspace{-0.8cm}
  \hspace{-1.6cm}
  Surprise$\rightarrow$Fear \vspace{-0.5cm}\\
 \end{tabular}
 % -------------------------------------------------------------------------
 \hrule
\vspace{0.2cm}
\caption{\small{Given source image and target emotion distribution, our system recolors the image by automatically selecting target images (from Emotion6 dataset \cite{Peng2015}) that are semantically similar to source image and their emotion distribution matches provided target emotion distribution. \textbf{Results are best visible in color print.}}\vspace{-0.3cm}}
\label{fig:transferImagesDemo}
\end{figure}
\end{center}

One of the important factors to judge the value of any multimedia content is it's emotional impact on the viewer. Images that evoke negative emotions tend to be more memorable \cite{khosla2015understanding} and online advertisement videos inducing surprise and joyous emotions have been shown to retain more viewers' concentration and attention \cite{teixeira2012emotion}. Unfortunately, most of the popular photo and video editing tools do not allow users to transform an image by indicating the emotional impact on it's viewers, as desired by the photographer. Fig.~(\ref{fig:transferImagesDemo}) shows few of the possible examples of such transformations.

\vspace{-0.3cm}
In this paper, we present a framework for image transformation such that the transformed source image has the affect, as desired by the user. 
Unlike previous methods \citep{kim2016image,Peng2015,xu2013image}, in our algorithm user has to input only the source image and seven dimensional discrete probability distribution representing ratio of seven emotions (anger, disgust, fear, joy, sadness, surprise and neutral). 
We use features extracted from top layers of deep convolutional networks (that encapsulate both in-image-context and content information) to select target images that have similar content and spatial context as the source image. In addition to that, we also make sure that emotion distribution of selected target images matches with the desired emotion distribution. To perform emotion transformation, low-level color features are extracted form these selected images and weighted combination of these features is then applied to the provided source image. \mFIG{\ref{fig:transferImagesDemo}} shows transformed images generated by our algorithm alongwith the automatically selected target images. The results of the performed user study shows that transformed images are more close to the desired emotion distributions (shown below target images) as compared to the corresponding source images.

Learning to generate a desired affect, either to change an image or to enhance it, will empower a photographer to manipulate the image according to the message he/she wants to convey or to align it with the content. For example, same photograph of scenic landscape can be manipulated to have gloomy feeling or joyous feeling of spring.
Similarly, our method could be used to edit the multimedia content for the VR and AR environments. 
 % --------------------------------------------------------------------------------------------------------------------------------------------------
% --------------------------------------------------------------------------------------------------------------------------------------------------
\section{Related Work}
\label{sec:litRe}

An important factor to judge the value of any multimedia content is the emotional impact it evokes. %Mostly the multimedia content is created to have specific emotional impact. 
%Emotional impact of the multimedia content plays vital role in how it is consumed. 
Music, lighting/illumination, camera pose and color etc.  have been used by movie directors to enhance the desired mood of the scene and its emotional impact on the viewers. For example, dark and gloomy colors are used to induce depression or fear, warm colors are used for positive emotions, fast moving camera is used to capture energy in the scene and tempo of the music is used to grab the attention of the viewer(s). 
Photographers on the other hand don't have all these tools and are restricted to manipulation of colors and textures only once the photograph has been taken.

\ignore{\subsection{Affective Analysis of Images}

\textcolor{red}{Colors define the affect of the image on the consumer. 
 There have been considerable study regarding the affect and color. 
Interpret-able features showed how the texture, edges and colors represent the affect. 
\\
Different emotion models have been proposed. 
MOHSEN: Talk about how different emotion models use Valance Arousal. This will be good place to introduce what they mean. 
}
}

\subsection{Colorization and Color/Style Transfer}
Image color transformation has been employed for variety of tasks, from colorization of the gray scale images to color de-rendering and image transfer. 
Where colorization of gray-scale images \cite{zhang2016colorful} and color de-rendering \cite{rushdi2013color} use learned models for the task (requiring only the input image); most of the image transfer algorithms require both source and target images as input.
Pouli and Reinhard \cite{pouli2010progressive} use the histogram matching algorithm to transfer the color and tone from the target image to the source image. \cite{hristova2015style} recolorizes the image such that style of the target image is transferred to the input image. Recently Gatys et al. \citep{Gatys2016ImageST} used deep convolutional network trained for object detection to transfer style of the target image to the source image. They claim that their algorithm understand how to separate the content and style because they use appropriate combination of high level and low-level features. However, their transformation is style transfer at the cost of realism. 

\subsection{Image Emotion Assignment}
Affective computing and emotion assignment to the image has been a popular topic in the last decade in the multimedia and computer graphics community. For brevity, we describe few of the proposed techniques in this section. Works like \citep{he2015image, murray2011towards} have tried to represent emotion by a concept and assigned each concept a color pallet designed on the basis of color emotion literature. On the other hand, example based approaches require a target image, representing desired affect, to be provided \citep{Peng2015}.
First technique described in \citep{he2015image, murray2011towards} suffers from the problem of modeling human perception that restricts it to few concepts and \citep{Peng2015} puts the burden on the user to provide an appropriate target image. Initially it looks an appropriate constraint to place on this under-constrained problem. However, finding an image which is both spatially similar to the input image and evokes emotion as user desires is difficult, ambiguity ridden and over-burdens the user. 

%He et al. \citep{he2015image} uses the predefined color emotion model to transform the color of the image.

Recent works have explored learning from data to understand the relationship between the images and different models of affect \cite{Peng2015, Machajdik2010}. Ali et al. \citep{affectHlc2017} have studied how high level concepts present in the images are related to the affect they induce. \cite{xue2013learning, xu2013image, kim2016image}  have used the datasets to learn and/or transfer the emotion to the input image. Xue et al. \cite{xue2013learning} model low-level color and tone features to represent different emotions of the movie clip clustered on the basis of genre and director of the film. Xu et al. \citep{xu2013image} trained Gaussian Mixture Model (GMM) on the super-pixels of the images that have been clustered into emotion-scene subgroups. The input image is first matched to a scene-subgroup in the desired emotion and then each super-pixel is transformed by minimizing energy equation that tries to find mapping from learned GMM to the super-pixels of input image. 
The above method uses low-level features and cluster their data into discrete classes of emotions. These methods are not suitable for our case where we have distribution of emotion and emotion-scene clustering is not feasible. 

Kim et al. \cite{kim2016image} instead of over-segmenting the input image use semantic segmentation. For each input-image-segment,  semantically compatible segments are searched from the database while minimizing on position, scale, lightness of the segment and closeness of user supplied Valance Arousal (VA) score to VA score of the image to whom selected segment belongs to. Due to their reliance on the semantic segmentation they constraint themselves to the landscape images, they also rely on VA scores which are not easily interpret-able by the humans and each segment's color is changed by just changing the mean value of the segment. 

For our experimental settings, we use the dataset introduced by Peng et al. \cite{Peng2015} where discrete emotion distribution is associated with each image. We use features extracted from the top layer of Deep CNN that capture both the content and spatial context information including the color and texture information at different locations of the image. We are not constrained by the semantic segmentation of an image and allow much more control of desired emotion by providing user ability to pick any discrete distribution of the emotions. Emotion distribution remains human interpret-able with the freedom to choose variety of combination of emotions.

\vspace{-0.4cm}
\section{Dataset}
\begin{figure}[h!]
\center
 \vspace{-0.1cm}
\includegraphics[width=0.25\linewidth]{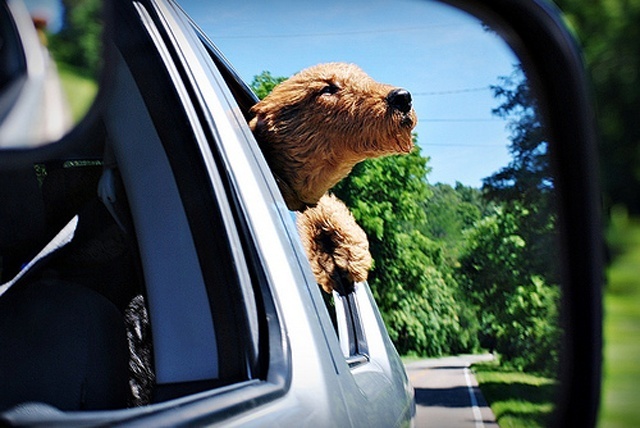} 
\hspace{0.5cm}
\includegraphics[width=0.30\linewidth]{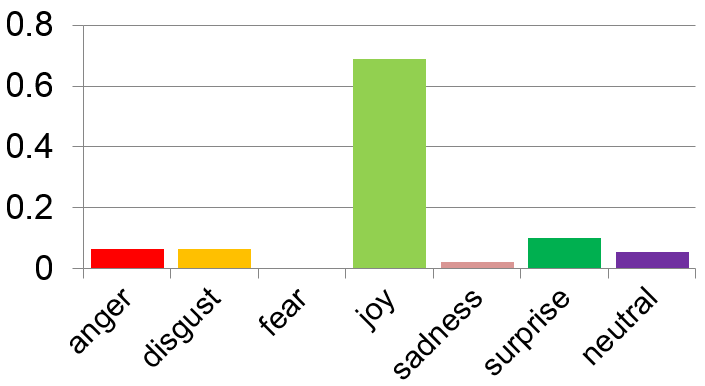} \\
\vspace{0.3cm}
\caption{\small{An example image from Emotion6 with its corresponding Emotion Distribution.}\vspace{-0.3cm}}\label{fig:emo6}
\end{figure}

\textbf{Emotion6} The results reported in this paper use a well known affective image dataset named Emotion6 \cite{Peng2015} that consists of 1980 images. Each image in Emotion6 dataset has an emotion distribution associated with it that indicates the probability of six Ekman's basic emotions \cite{ekman1992} and a neutral one being evoked in its viewers. The \mFIG{\ref{fig:emo6}} shows an example image from Emotion6 dataset with its corresponding emotion distribution. The higher the probability of an emotion, there are more chances for particular emotion being evoked in its viewers. \\

\textbf{ArtPhoto} A subset of images from Artphoto dataset~\cite{Machajdik2010} are also used in our experiments. In ArtPhoto dataset, images are carefully taken by artists to induce specific emotion in their viewers by keenly selecting interpretable aesthetic features related to image composition, colors and lighting effects, etc.

\section{Methodology}
We propose a novel method that unlike previous methods (e.g. \cite{Peng2015}) requires only a source image and a target emotion distribution for affective image transformation. By eliminating the need for the suitable target image, we free the user from the burden of searching it in order to obtain the transformed image that could elicit desired emotion distribution. Semantic information of source image is taken in consideration while performing emotion transformation to ensure that color assigned to the objects present in the image lies in the color space in which those objects naturally exist. This is also crucial for the naturalness of the transformed image, as blue sky or blue sea might induce pleasant feelings in its viewers but blue tree would look unnatural and therefore may not elicit pleasantness in the observers.

\begin{center}
\begin{figure*}[t!]
 \centerline{\includegraphics[width=0.9\linewidth]{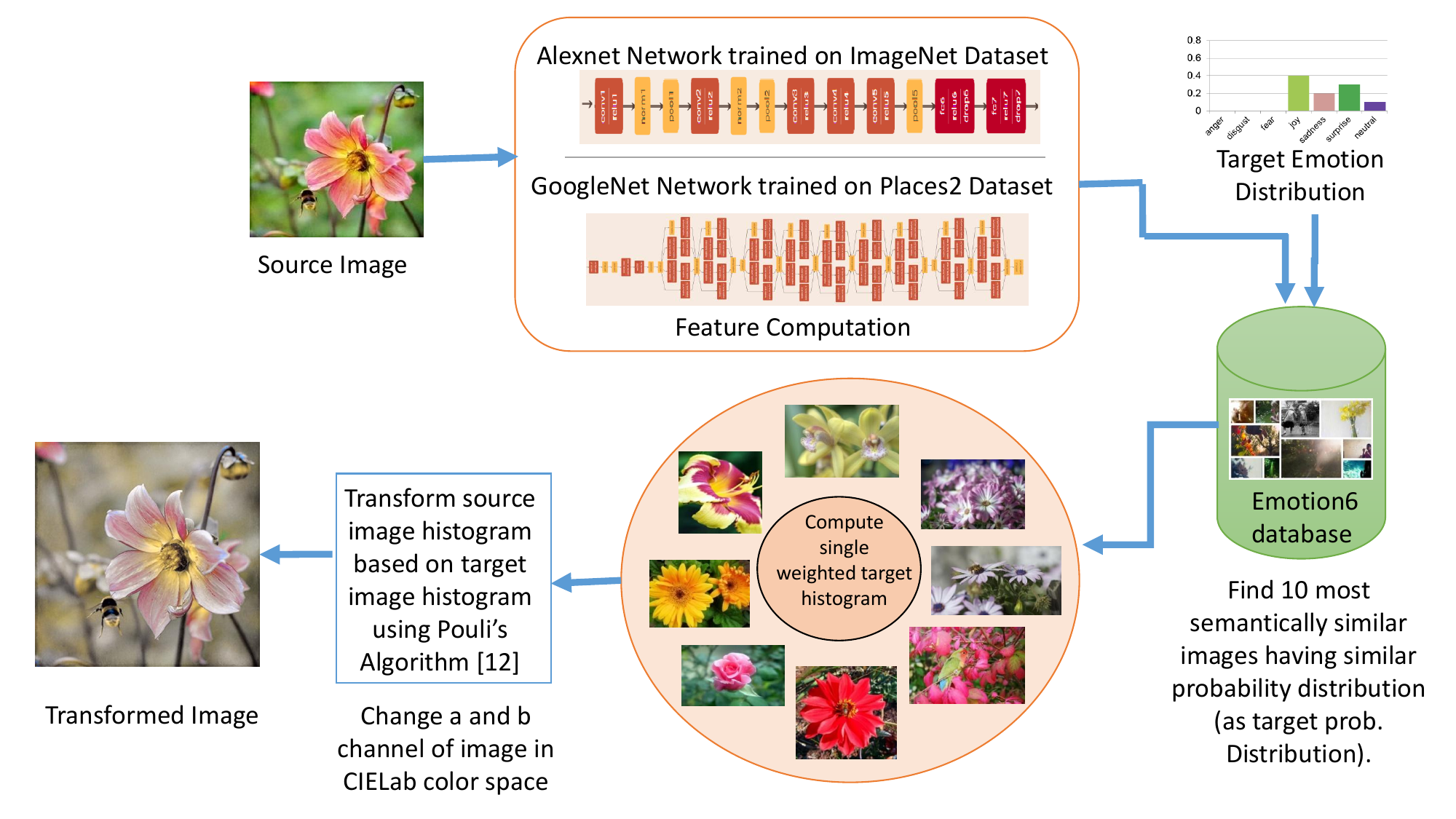}}
\caption{Block diagram of proposed Affect assignment algorithm. The AlexNet and GoogleNet network visualizations are created using \cite{wiki:xx}. \vspace{-0.6cm}}
\label{fig:res}
\end{figure*}
\end{center}

\begin{center}
\begin{figure}[b!]
\centering{
\includegraphics[width=0.7\linewidth]{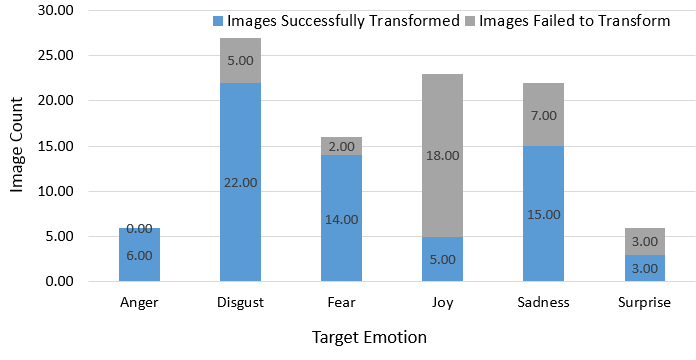}}\label{fig:stats}
\caption{\small{The number of images being transformed towards each emotion.}}\label{fig:taggingStats}
\end{figure}
\vspace{-0.55cm}
\end{center}
%-------------------------------------------------------------------------------
\vspace{-0.8cm}
The block diagram of our emotion transfer algorithm is shown in \mFIG{\ref{fig:res}} where a user is requested to input only a source image $I_s$ and the desired emotion distribution $P_d$. 
Similar to \cite{Peng2015}, in our method color tone is adjusted using the algorithm proposed in \cite{pouli2010progressive} that relies on the color histogram of the target image to perform the transformation. However, we don't rely on any user input targeted image for the calculation of target color histogram. 

Let, $D$ be database constructed from image-emotion distribution pair.
Currently, we are only using Emotion6 dataset as our database since it is the only dataset where each image $I$ is associated with an emotion probability distribution $\bP$ = $\left\{\text{$p^1$,$p^2$,...,$p^7$}\right\}$ representing probability distribution of Ekman's six basic emotions and a neutral one.

Given an input image $I_s$ and the desired emotion distribution $P_d$, we construct a histogram $\bh$ as weighted summation of histograms of target images in the database $D$; $h = \sum_{I_i \in T} ( w_i * \mathbf{h}_i )$
%\begin{equation} \label{eq:5_1}
%h = \sum_{I_i \in T} ( w_i * \mathbf{h}_i )
%\end{equation}

where $w_i$ is the assigned weight to $I_i$,  $|T|=K$ and $T\subseteq D$. $T$ is subset of $K$ images such that $d=\sum_{I_i \in T} distance(P_d,P_i)+ distance(I_i,I_s)$ is minimized. To minimize $d$ we solve the problem in two steps.
First we construct subset $D^\ast \subseteq D$ by selecting only those images $I_i \in D$ having emotion distribution similar to the input emotion distribution. 
Bhattacharyya coefficient \cite{wiki:xxx} is used as similarity measure between emotion distributions
\begin{equation}  \label{eq:1}
bc_i = distance(P_d,P_i) = \sum_{k=1}^{7} \sqrt{P_d(k) \cdot P_i(k)}
\end{equation}

%\begin{equation}  \label{eq:1}
%bc_i = distance(P_d,P_i) = \sum_{k=1}^{7} \sqrt{P_d(k) \cdot P_i(k)} \qquad \forall i \in \{1, ..., N\}
%\end{equation}

%of all $N$ images in $D$ and $P_d$, eq.~(\ref{eq:1}),
To find the required target image we search our database $D$ and instead of selecting just one target image, we select subset of images $D^\ast$ that has emotion distribution similar to the desired emotion distribution, as described in  eq.~(\ref{eq:2}).

%Bhattacharyya coefficient \cite{wiki:xxx} is computed between emotion distribution of all $N$ images in $D$ and $P_d$, eq.~(\ref{eq:1}), and the images that satisfy eq.~(\ref{eq:2}) are selected.
\begin{equation}  \label{eq:2}
D^\ast = \{\text{$I_i$ : $bc_i$ > $\omega$}\},       \hspace{1cm}    where \hspace{0.5cm} \omega = 1.5 * mean (bc_{i=\{1,2,...,N\}})
\end{equation} 
%----------------------------------------------------------------------------

%using the K-nearest neighbor algorithm (see eq.~(\ref{eq:4})).
To compute similarity between the images $I_s$ and $I_i \in D_{\ast}$ we consider not only similar concepts that two images have in common but also the spatial and color combinations in which these concepts exist. In order to get target images that are semantically more close to the source image we used top layers of a Convolutional Neural Network (CNN) as a recent study \cite{zeiler2014visualizing} shows that these layers capture high level concepts (related to in-image context and semantics) along with discriminant low level features. We used $fc7$ and $pool5$ layers of AlexNet \cite{Krizhevsky2012NIPSImageNet} and GoogleNet networks \cite{zhou2014Places}, trained on ImageNet and Places2 datasets respectively, to compute these features.
%Depending upon the datasets, both networks are trained on, the fc7 and pool5 features encode information corresponding to the objects and the scenes present in an image.
For each image, AlexNet and GoogleNet models give us 4096 and 1024 dimensional output vectors which are then normalized and concatenated together to get a single 5120 dimensional vector denoted as $\mathbf{f}$. 

We apply K-nearest-neighborhood algorithm to select set of $K$ target images $T$ = $\{I_{t_k}\}_{k=1}^K$ from $D^\ast$ that are semantically closer to the provided source image, using $\mathbf{f}$, eq.(\ref{eq:4}). We chose $K$ to be 10 in our experiments because mean distance (representing semantic gap) between source and nearest neighborhood images increases drastically as we further increase K and implications of which is that our results deteriorate qualitatively.

\begin{equation} \label{eq:4}
\underset{T} {\mathrm{argmin}} ~\| \text{$\mathbf{f}_{s} - \mathbf{f}_{j}$} \| \qquad \forall j \in D^\ast
\end{equation}

where $\mathbf{f}_{s}$ represents the source image features. 
For each selected target image $i$ in $K$, a histogram $\bh_i$ is computed as in \cite{pouli2010progressive} and we take weighted summation using the Bhattacharyya coefficient $bc_i$ of $K$ selected images $h = \sum_{i=1}^{K} ( bc_i * \mathbf{h}_i )$.
%\begin{equation} \label{eq:5}
%h = \sum_{i=1}^{K} ( bc_i * \mathbf{h}_i )
%\end{equation}
 Note that, $bc_i$ measures similarity between target emotion distribution and the selected images. In this way the target histogram is more affected by the image whose distribution is more closer to the target emotion distribution. After computing target image histogram, we used color transfer algorithm in \cite{pouli2010progressive} to compute the final transformed image.

%------------------------------------------------------------------------------
\begin{center}
\begin{figure*}[h!]
\center
 \begin{tabular}{m{1.2cm}m{2cm} m{2.5cm} m{3.0cm} m{2.3cm} } 
 \hline
\small{\textbf{Target Emotion}} & \small{\textbf{Source Image}} &  \small{\textbf{Target Emotion Distribution}} & \small{\textbf{{Selected Images from Emotion6 Dataset}}} & \small{\textbf{{Transformed Image}}}\\ [0.5ex] 
 \hline
 \end{tabular}
  % -------------------------------------------------------------------------
 \begin{tabular}{m{0.8cm} m{2.0cm} m{2.5cm} m{3.0cm} m{2.0cm} } 
 \small{ Anger } &
 \vspace{-0.2cm}
  \subfloat{ \includegraphics[width=0.9\linewidth]{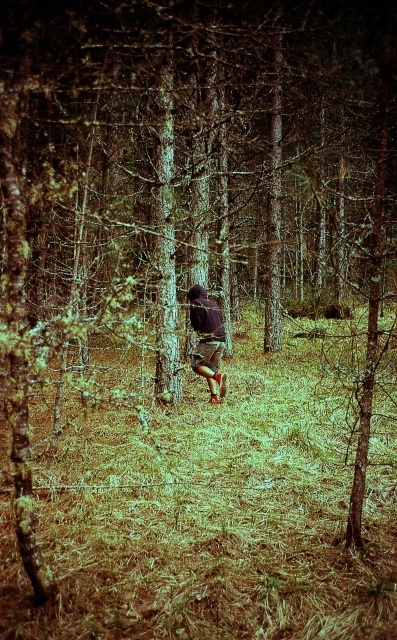}} &
   \vspace{-0.4cm} 
 \subfloat{\includegraphics[height=2.3cm, width=2.5cm]{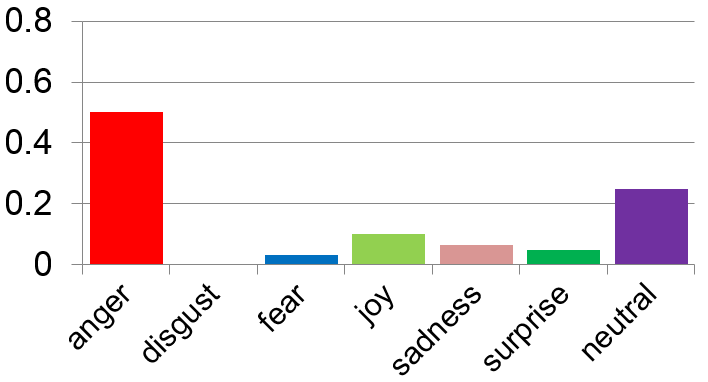}} &
 \vspace{-0.2cm}
 \subfloat{ \includegraphics[width=1\linewidth]{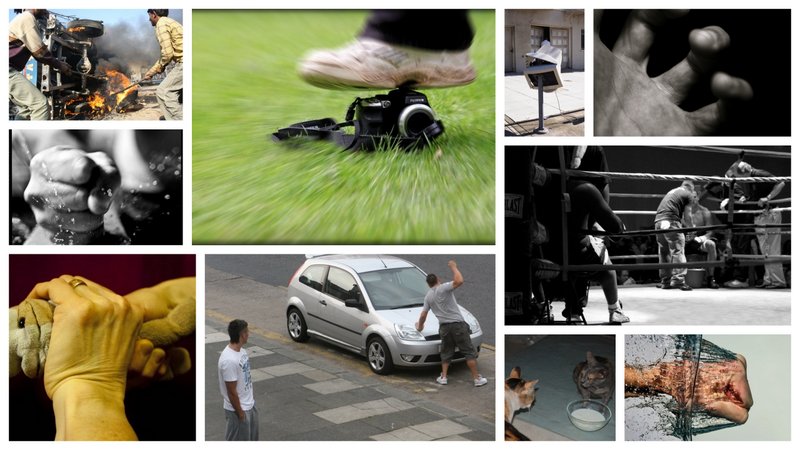}} &
 \vspace{-0.2cm}
 \subfloat{ \includegraphics[width=0.9\linewidth]{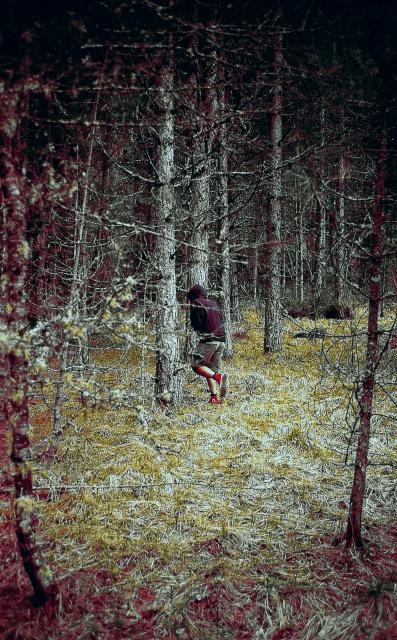}} 
 \end{tabular}
   % -------------------------------------------------------------------------
 \begin{tabular}{m{0.8cm} m{2.0cm} m{2.5cm} m{3.0cm} m{2.0cm} } 
 \small{ Disgust } &
 \vspace{-0.2cm}
  \subfloat{ \includegraphics[width=\textwidth, width=0.9\linewidth]{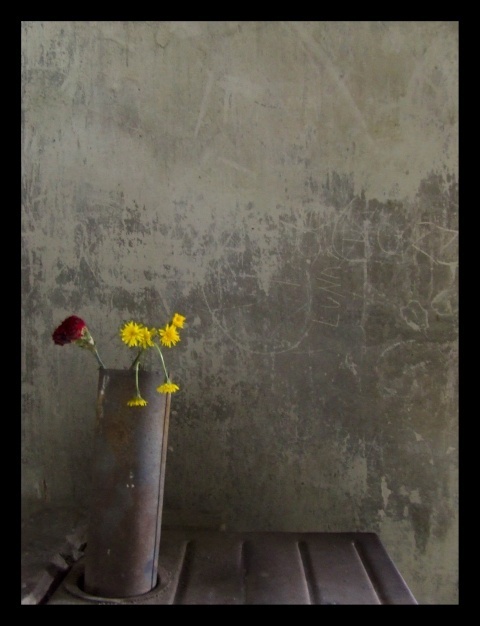}} &
   \vspace{-0.4cm} 
 \center \subfloat{\includegraphics[width=\textwidth, height=2.3cm, width=2.5cm]{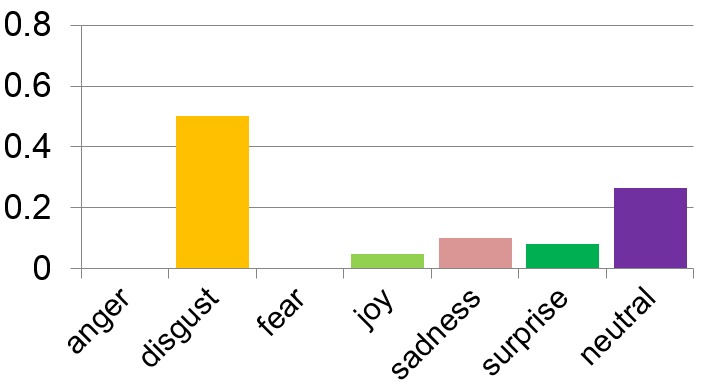}} &
 \vspace{-0.2cm}
 \subfloat{ \includegraphics[width=\textwidth, width=1\linewidth]{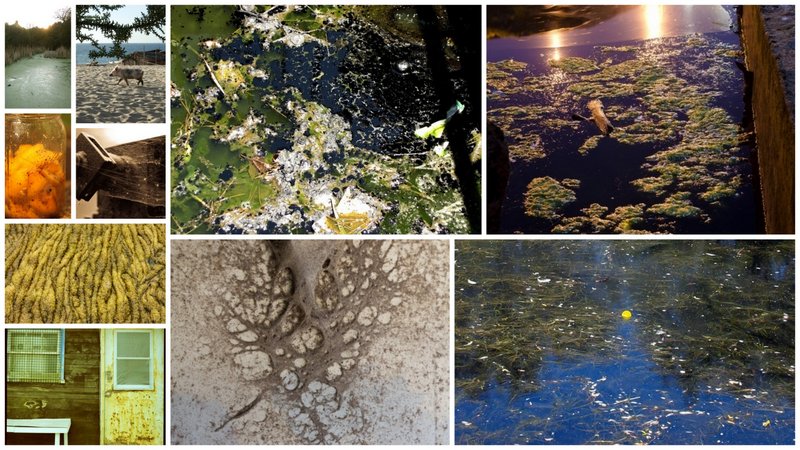}} &
 \vspace{-0.2cm}
 \subfloat{ \includegraphics[width=\textwidth, width=0.9\linewidth]{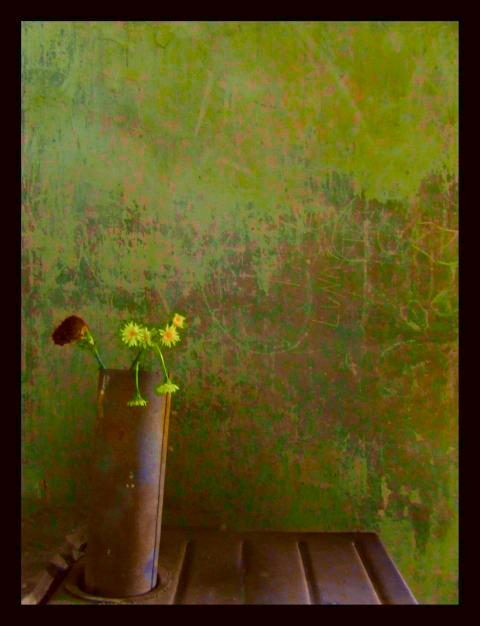}} 
 \end{tabular} 
% -------------------------------------------------------------------------
 \begin{tabular}{m{0.8cm} m{2.0cm} m{2.5cm} m{3.0cm} m{2.0cm} } 
 \small{ Fear } &
 \vspace{-0.2cm}
  \subfloat{ \includegraphics[width=\textwidth, width=0.9\linewidth]{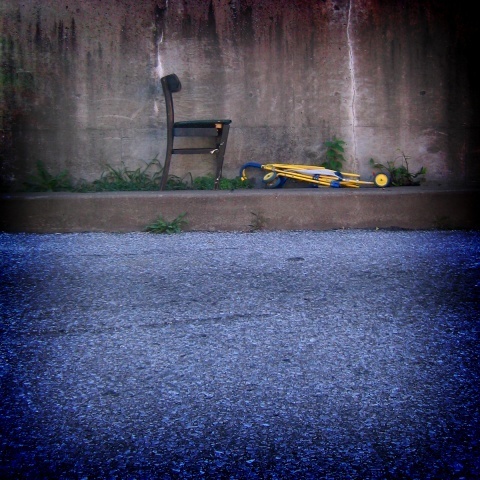}} &
   \vspace{-0.4cm} 
 \center \subfloat{\includegraphics[width=\textwidth, height=2.3cm, width=2.5cm]{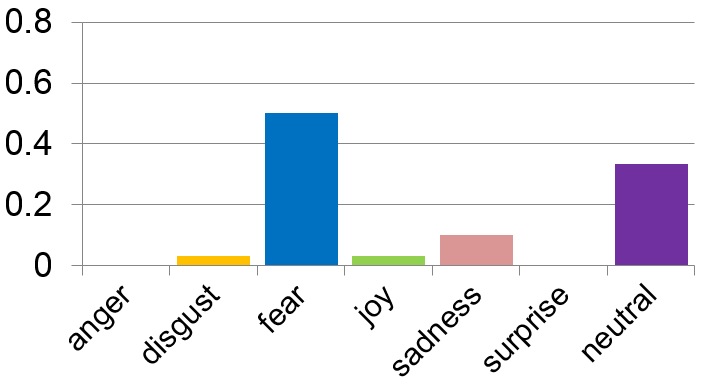}} &
 \vspace{-0.2cm}
 \subfloat{ \includegraphics[width=\textwidth, width=1\linewidth]{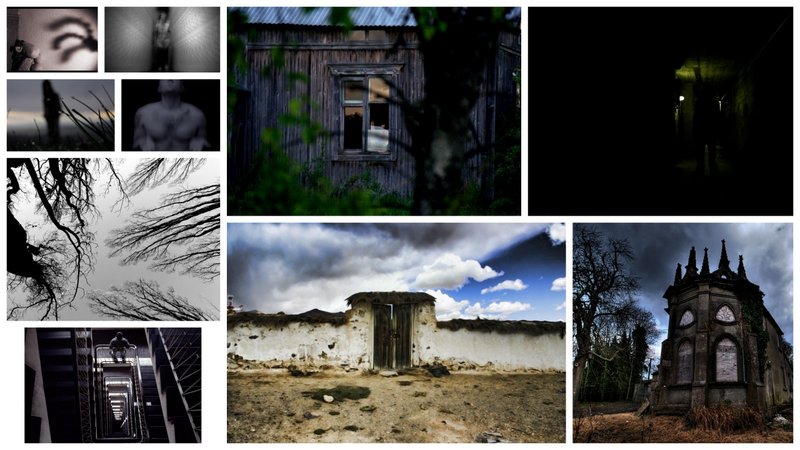}} &
 \vspace{-0.2cm}
 \subfloat{ \includegraphics[width=\textwidth, width=0.9\linewidth]{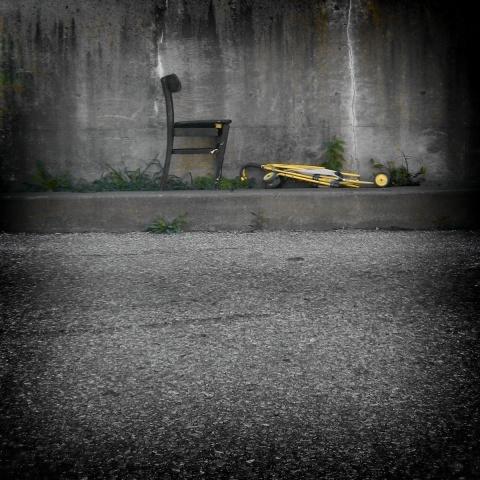}} 
 \end{tabular}
% ------------------------------------------------------------------------
 \begin{tabular}{m{0.8cm} m{2.0cm} m{2.5cm} m{3.0cm} m{2.0cm} } 
 \small{ Joy } & 
 \vspace{-0.2cm}
  \subfloat{ \includegraphics[width=\textwidth, width=0.9\linewidth]{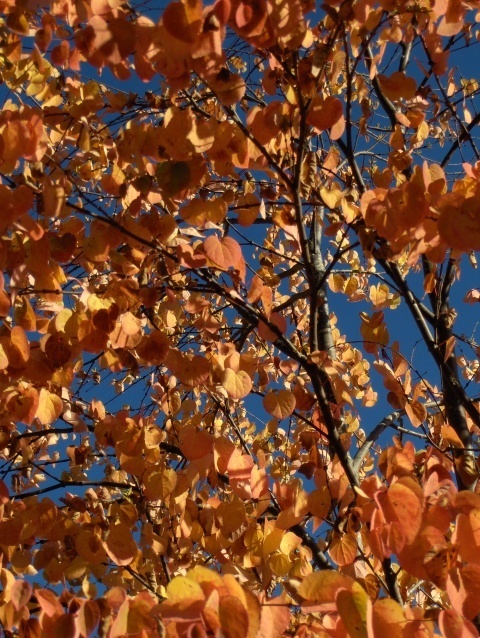}} &
   \vspace{-0.4cm} 
 \center \subfloat{\includegraphics[width=\textwidth, height=2.3cm, width=2.5cm]{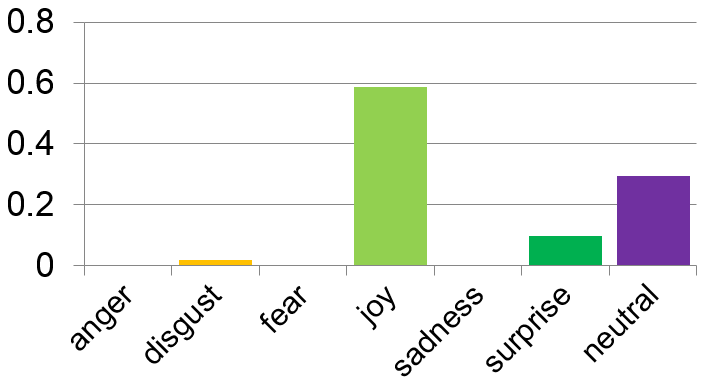}} &
 \vspace{-0.2cm}
 \subfloat{ \includegraphics[width=\textwidth, width=1\linewidth]{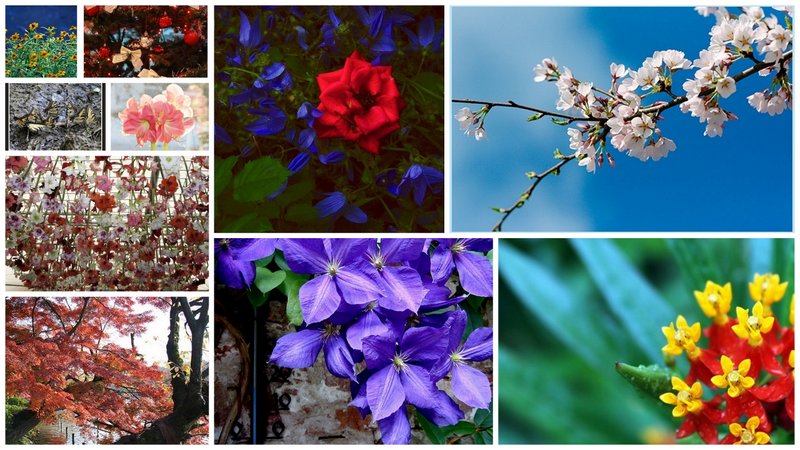}} &
 \vspace{-0.2cm}
 \subfloat{ \includegraphics[width=\textwidth, width=0.9\linewidth]{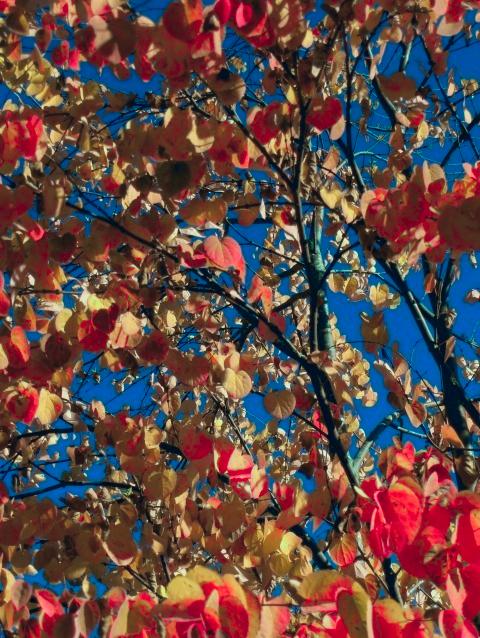}} 
 \end{tabular}
  % -------------------------------------------------------------------------
 
   % -------------------------------------------------------------------------
 \begin{tabular}{m{0.8cm} m{2.0cm} m{2.5cm} m{3.0cm} m{2.0cm} } 
 \small{ Sadness } &
 \vspace{-0.2cm}
  \subfloat{ \includegraphics[width=\textwidth, width=0.9\linewidth]{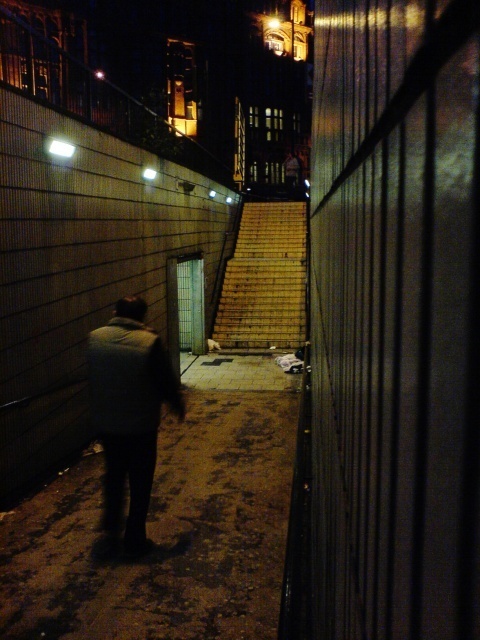}} &
   \vspace{-0.4cm} 
 \center \subfloat{\includegraphics[width=\textwidth, height=2.3cm, width=2.5cm]{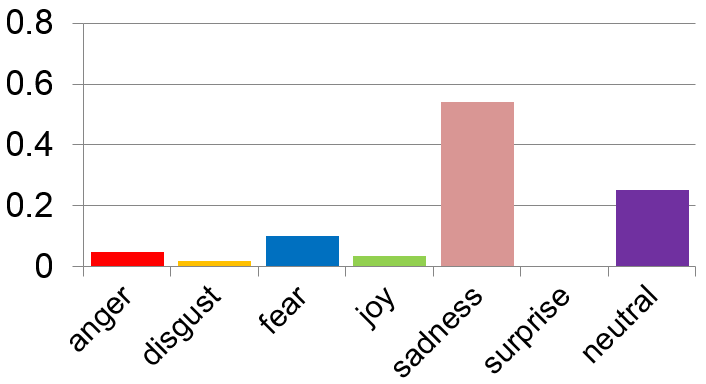}} &
 \vspace{-0.2cm}
 \subfloat{ \includegraphics[width=\textwidth, width=1\linewidth]{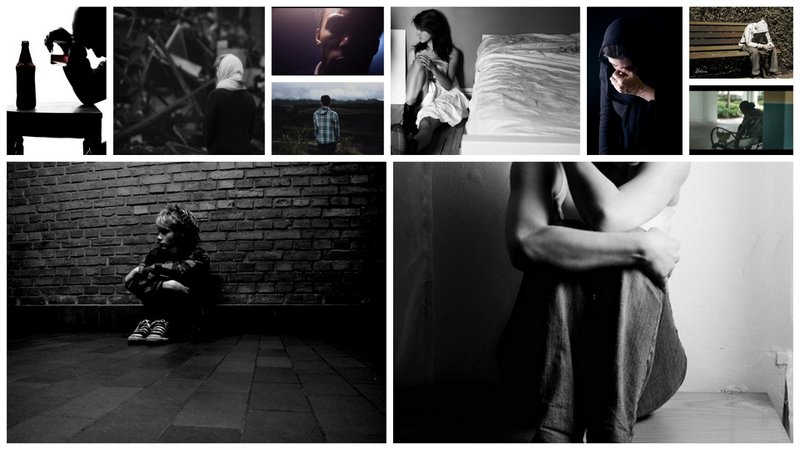}} &
 \vspace{-0.2cm}
 \subfloat{ \includegraphics[width=\textwidth, width=0.9\linewidth]{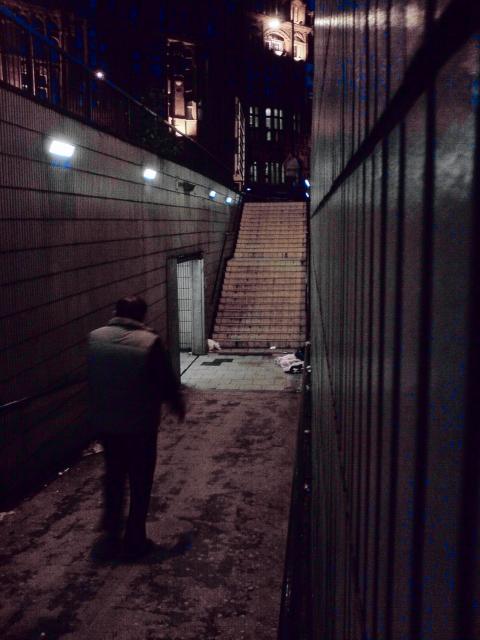}} 
 \end{tabular}
 \vspace{0.2cm}
\caption{\small{Examples of transferring evoked emotion distribution. Second column contains the source image. Third and fourth column indicates the Target Emotion Distribution provided by the user and the images selected by our algorithm respectively.}\vspace{-0.6cm}}
\label{fig:transferImagesFinal}
\end{figure*}
\end{center}
%-----------------------------------------------------------------------------------
\vspace{-1cm}
\section{Experiments and Results}
\begin{center}
\begin{figure*}[t!]
\center
\hrule
 \begin{tabular}{m{1.55cm} m{2.35cm} m{1.45cm} : m{1.55cm} m{2.35cm} m{1.45cm}} 
\small{\textbf{Input Image}} & \small{\textbf{{Target Images}}} & \small{\textbf{{Output Image}}} &
\small{\textbf{Input Image}} & \small{\textbf{{Target Images}}} & \small{\textbf{{Output Image}}} \\ [0.5ex] 
 \end{tabular}
  \hrule
 
  % -------------------------------------------------------------------------

 \begin{tabular}{m{1.55cm} m{2.35cm} m{1.55cm}  : m{1.55cm} m{2.35cm} m{1.55cm}} 
 \vspace{-0.2cm}
  \subfloat{ \includegraphics[width=\textwidth ,width=1\linewidth]{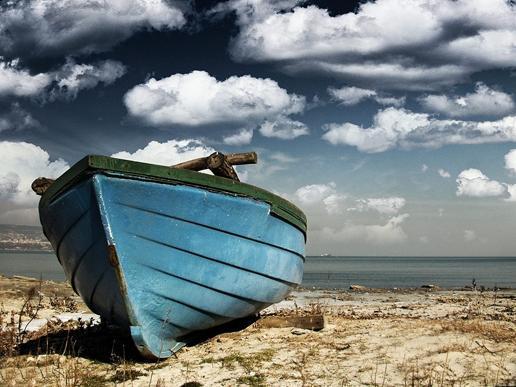}} &
 \vspace{-0.2cm}
 \hspace{-0.3cm}
 \subfloat{ \includegraphics[width=\textwidth, width=1\linewidth]{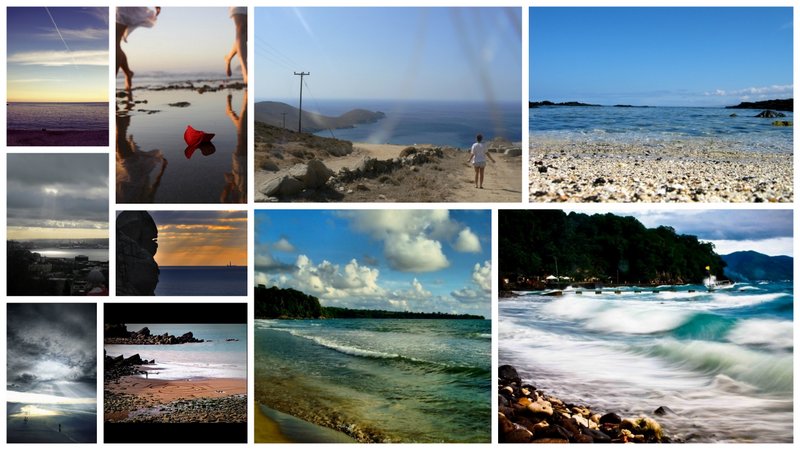}} &
 \vspace{-0.2cm}
 \hspace{-0.5cm}
 \subfloat{ \includegraphics[width=\textwidth, width=1\linewidth]{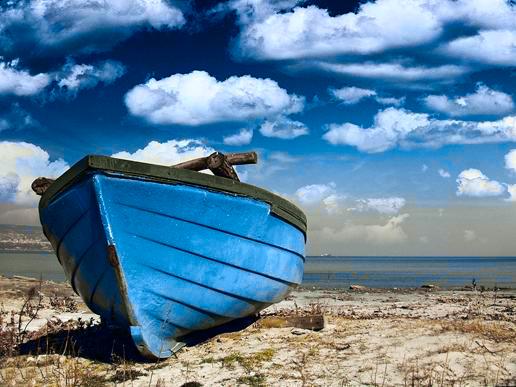}} &
  \vspace{-0.2cm}
  \subfloat{ \includegraphics[width=\textwidth ,width=1\linewidth]{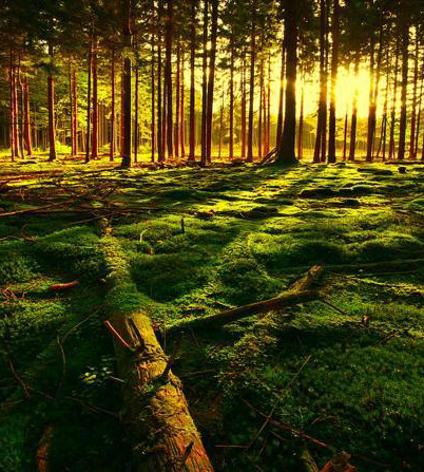}} &
 \vspace{-0.2cm}
  \hspace{-0.3cm}
 \subfloat{ \includegraphics[width=\textwidth, width=1\linewidth]{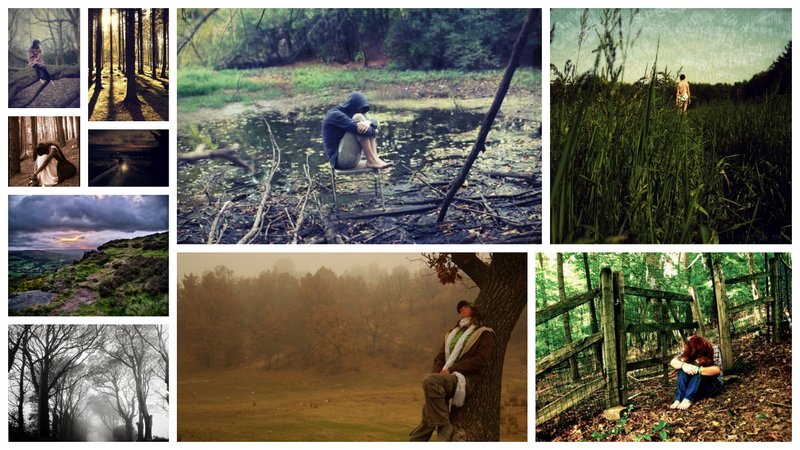}} &
 \vspace{-0.2cm}
  \hspace{-0.5cm}
 \subfloat{ \includegraphics[width=\textwidth, width=1\linewidth]{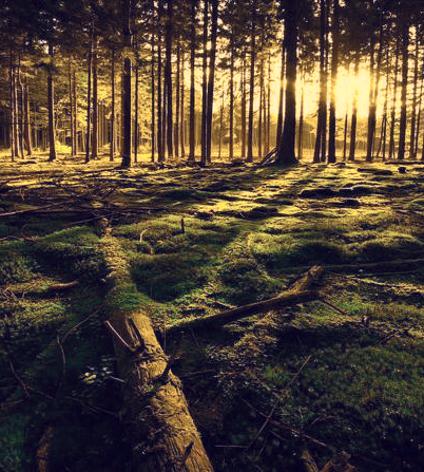}} \\
  \vspace{-0.2cm} 
  \small{ Sad} &  & \small{ Joy} & \small{ Excitement} &  & \small{ Sadness} \\
 
 \end{tabular}
 
 % -------------------------------------------------------------------------

  \begin{tabular}{m{1.55cm} m{2.35cm} m{1.55cm} : m{1.55cm} m{2.35cm} m{1.55cm}} 
 \vspace{-0.2cm}
  \subfloat{ \includegraphics[width=\textwidth ,width=0.9\linewidth]{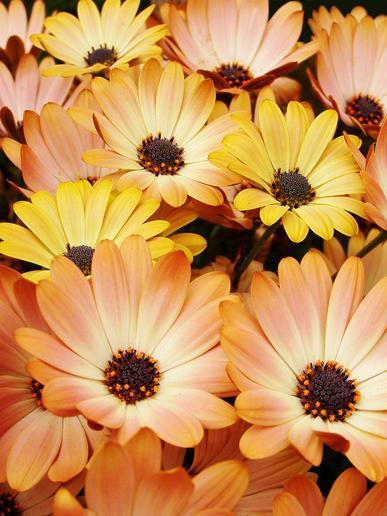}} &
 \vspace{-0.2cm}
  \hspace{-0.3cm}
 \subfloat{ \includegraphics[width=\textwidth, width=1\linewidth]{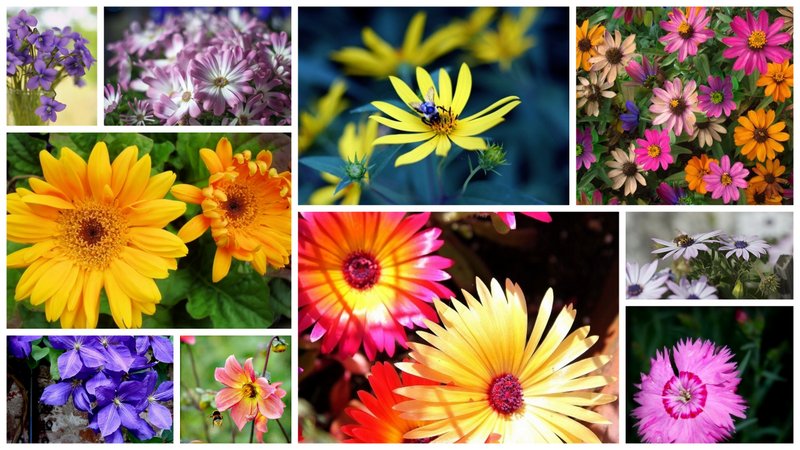}} &
 \vspace{-0.2cm}
  \hspace{-0.5cm}
 \subfloat{ \includegraphics[width=\textwidth, width=0.9\linewidth]{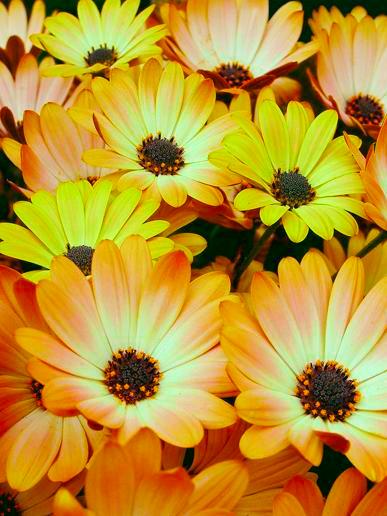}} &
  \vspace{-0.2cm}
  \subfloat{ \includegraphics[width=\textwidth ,width=0.9\linewidth]{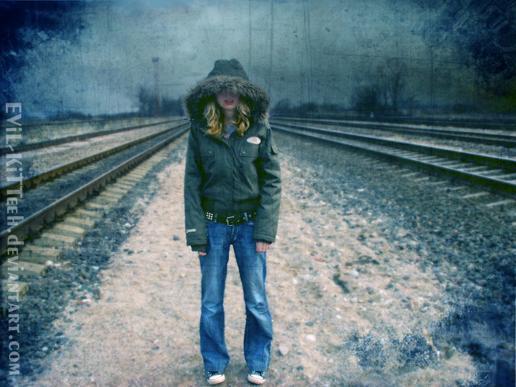}} &
 \vspace{-0.2cm}
  \hspace{-0.3cm}
 \subfloat{ \includegraphics[width=\textwidth, width=1\linewidth]{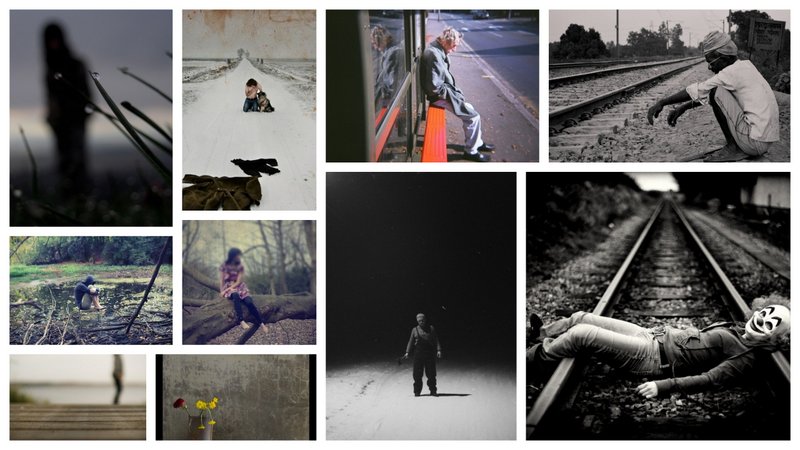}} &
 \vspace{-0.2cm}
  \hspace{-0.5cm}
 \subfloat{ \includegraphics[width=\textwidth, width=0.9\linewidth]{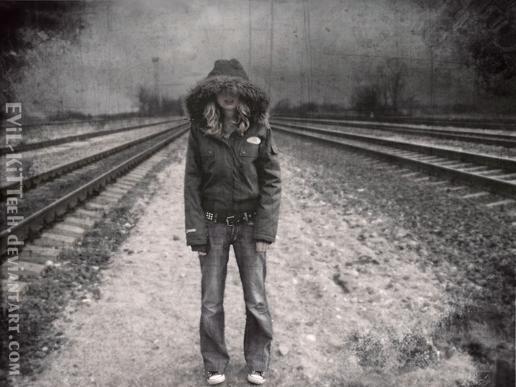}} 
   \vspace{-0.2cm} \\
  \small{ Excitement} &  &  \hspace{-0.5cm}\small{ Excitement Increased} & \small{  Sadness} &  &  \hspace{-1cm} \small{ Sadness Increased} \\
 \end{tabular}
 \vspace{0.2cm}
 % -------------------------------------------------------------------------
 \caption{\small{Emotion transformation on some images from ArtPhoto dataset. First row: change of affect class. Second row: enhancing the arousal value of induced emotion. }\vspace{-0.5cm}}\label{fig:emotionTransArtPhoto}
\end{figure*}
\end{center}

\begin{center}
\begin{figure*}[h!]
\center
 \begin{tabular}{m{1.2cm}m{2.0cm} m{2.5cm} m{3.0cm} m{2.3cm} } 
 \hline
\small{\textbf{Target Emotion}} & \small{\textbf{Source Image}} &  \small{\textbf{Target Emotion Distribution}} & \small{\textbf{{Selected Images from Emotion6 Dataset}}} & \small{\textbf{{Transformed Image}}}\\ [0.5ex] 
 \hline
 \end{tabular}
 
  % -------------------------------------------------------------------------

 \begin{tabular}{m{0.8cm} m{2.3cm} m{2.5cm} m{3.0cm} m{2.3cm} } 
 \small{ Anger } &
 \vspace{-0.2cm}
  \subfloat{ \includegraphics[width=\textwidth ,width=0.9\linewidth]{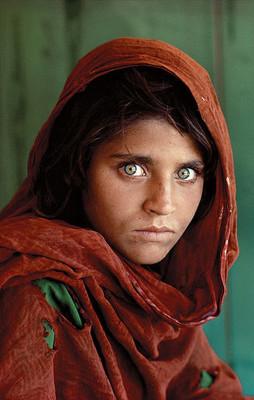}} &
   \vspace{-0.4cm} 
 \center \subfloat{\includegraphics[width=\textwidth, height=2.3cm, width=2.5cm]{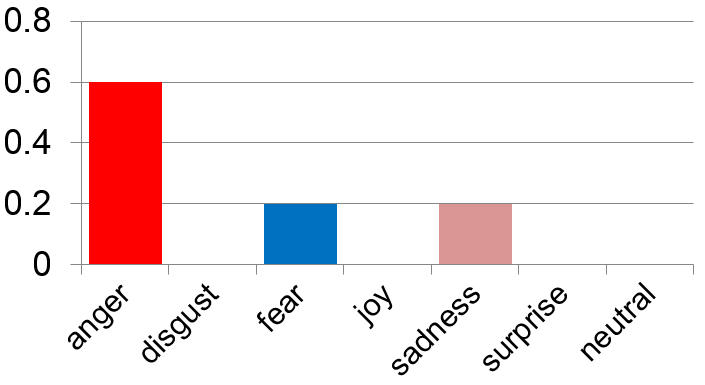}} &
 \vspace{-0.2cm}
 \subfloat{ \includegraphics[width=\textwidth, width=1\linewidth]{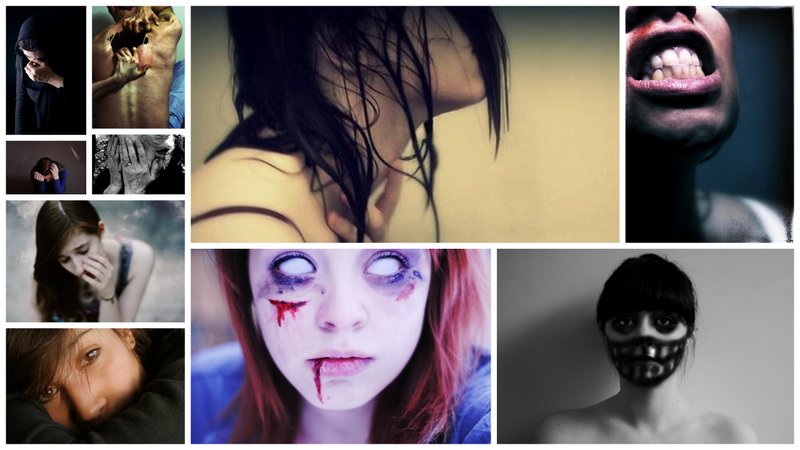}} &
 \vspace{-0.2cm}
 \subfloat{ \includegraphics[width=\textwidth, width=0.9\linewidth]{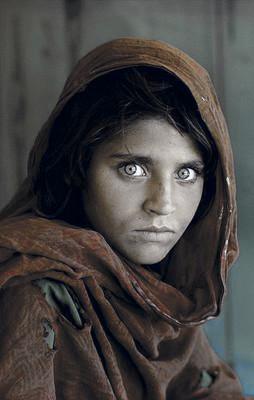}} 
 \end{tabular}
 
 % -------------------------------------------------------------------------

 \begin{tabular}{m{0.8cm} m{2.3cm} m{2.5cm} m{3.0cm} m{2.3cm} }
 \small{ Fear } &
 \vspace{-0.2cm}
  \subfloat{ \includegraphics[width=\textwidth ,width=0.9\linewidth]{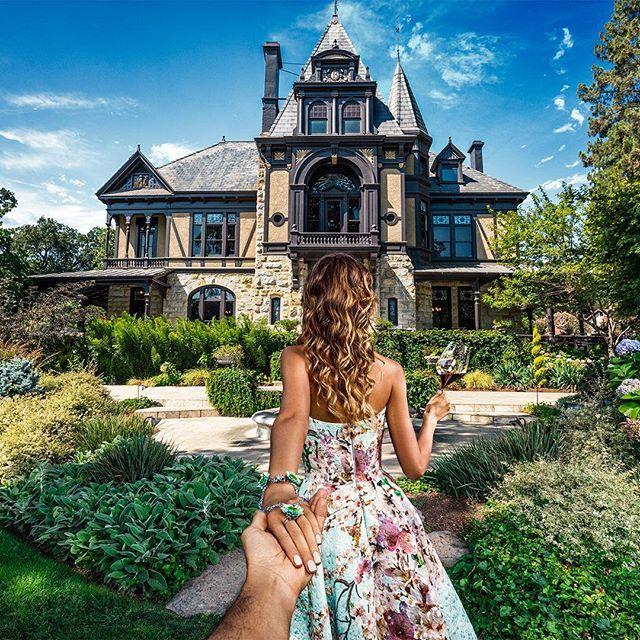}} &
   \vspace{-0.4cm} 
 \center \subfloat{\includegraphics[width=\textwidth, height=2.3cm, width=2.5cm]{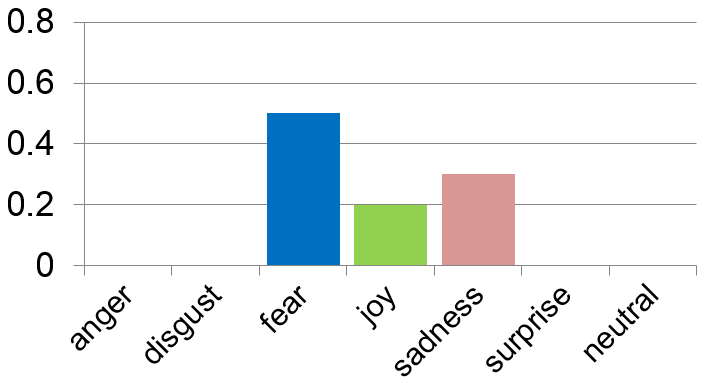}} &
 \vspace{-0.2cm}
 \subfloat{ \includegraphics[width=\textwidth, width=1\linewidth]{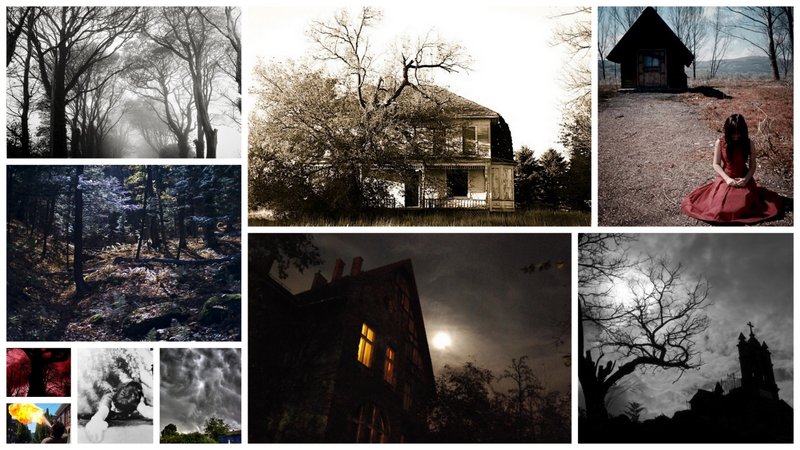}} &
 \vspace{-0.2cm}
 \subfloat{ \includegraphics[width=\textwidth, width=0.9\linewidth]{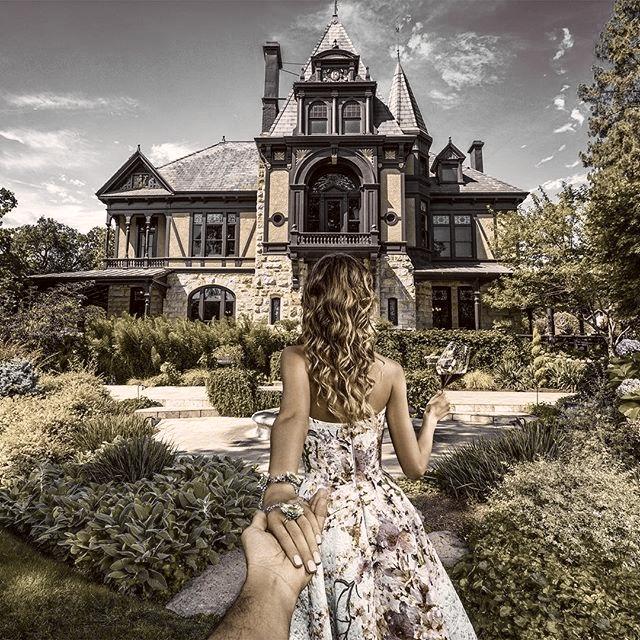}} 
 \end{tabular}
 \vspace{0.2cm} 
\caption{\small{Emotion transformation on some popular photographs. In the first row, we have shown to increase anger with bit of sadness and fear, whereas, in the second image we have tried to increase the element of fear with a bit of sadness.}}\label{fig:emotionTransFamous}
\end{figure*}
\end{center}

We apply our algorithm on images in Emotion6 dataset and few of the famous images. 
\mFIG{\ref{fig:transferImagesDemo}}, \mFIG{\ref{fig:transferImagesFinal}}, \mFIG{\ref{fig:emotionTransArtPhoto}} and \mFIG{\ref{fig:emotionTransFamous}} demonstrate our emotion transfer results. Considering that emotions are subjective in nature, we conduct a user study to evaluate how good our transformed image represent the targeted emotion. The emotion with highest value in provided emotion distribution is chosen to be target emotion.  

\textbf{Experiment 1} In our first experiment, we select top 100 neutral images, based on the ground truth, from Emotion6 dataset \cite{Peng2015} as source images. For each image, we use target emotion to be the one that selects target images with most similar probability distribution as the desired distribution. This is performed by computing Bhattacharyya coefficient (see equation \ref{eq:1}) between desired distribution and emotion distribution of selected target images by our algorithm.  The target emotion distribution is then obtained by setting the target emotion to 1 in the source image's emotion distribution and then normalizing the distribution such that it sums to 1. Out of the total 100 images, the number of images transformed for each emotion are shown in \mFIG{\ref{fig:taggingStats}}. A few source-transformed pairs of this user study are shown in \mFIG{\ref{fig:transferImagesFinal}}. To evaluate that the emotion distribution of transformed image has higher value for targeted emotion than the neutral, we presented these 100 source-transformed image pairs in random order to different subjects and ask them to tag the one image that induce more target emotion in them. User study was conducted through an online web portal that collects subject's responses to the shown images. The link was shared online, and undergraduate and graduate students of computer science department were requested to take part in the study.The demographics of the user study, therefore, consists of both male and female subjects, of age ranging from 17 to 25. We got a total of 1671 responses, where on average each image was tagged almost by 16 times. The statistics generated by this user study show that 65.0\% of the times our emotion transfer algorithm successfully transformed neutral images towards the target emotion.\ignore{, roughly comparable to 76.50\% reported in \cite{Peng2015} where they only used best transformations against each image to evaluate the emotion transfer algorithm.}

\mFIG{\ref{fig:taggingStats}} shows that some of our neutral images failed to transform to joy emotion, this is because the images chosen contain the neural objects and simple color transformation cannot make these images elicit joyous feeling. An example of such transformation is given in first row and second column of figure \ref{fig:emotionTransFail}. 
These results clearly demonstrate that high level concepts present in the images constraint how much the emotion of the image can be modified.

\textbf{Experiment 2} In the second experiment we transformed emotion of a subset of images from ArtPhoto dataset. The transformation results of few selected images are shown in \mFIG{\ref{fig:emotionTransArtPhoto}}. These results clearly depict that we are able to successfully perform emotion transformation on these images. In addition to images from ArtPhoto dataset, we have also tried to transform induced emotion of few popular photographs. The results of these images are shown in \mFIG{\ref{fig:emotionTransFamous}}. For all images in \mFIG{\ref{fig:emotionTransArtPhoto}} and \mFIG{\ref{fig:emotionTransFamous}} majority of subjects voted that transformed images are more close to the target emotion as compared to the source images.

 % -------------------------------------------------------------------------

%--------------------------------------------------------------------------------

%-----------------------------------------------------------------------------

\vspace{-0.8cm}
\begin{center}
\begin{figure*}[h!]
\center
\hrule
 \begin{tabular}{m{1.55cm} m{2.35cm} m{1.45cm} : m{1.55cm} m{2.35cm} m{1.45cm}} 
\small{\textbf{Input}} & \small{\textbf{{Target Images}}} & \small{\textbf{{Output}}} &
\small{\textbf{Input}} & \small{\textbf{{Target Images}}} & \small{\textbf{{Output}}} \\ [0.5ex] 
 \end{tabular}
  \hrule
 
  % -------------------------------------------------------------------------
 \begin{tabular}{m{1.55cm} m{2.35cm} m{1.55cm} : m{1.55cm} m{2.35cm} m{1.55cm}} 
 \vspace{-0.2cm}
  \subfloat{ \includegraphics[width=\textwidth ,width=1\linewidth]{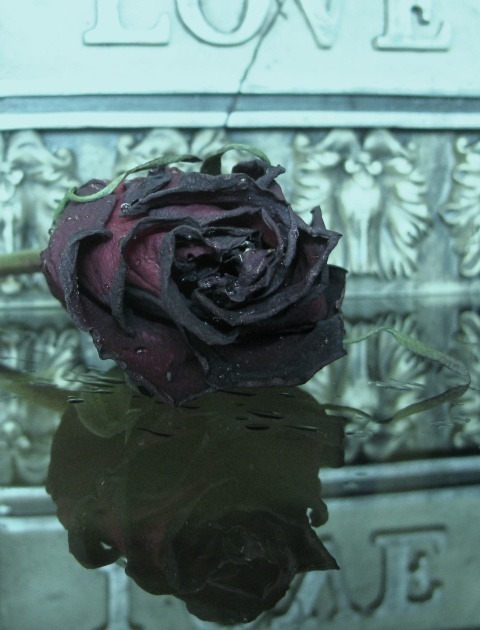}} &
 \vspace{-0.2cm}
 \hspace{-0.3cm}
 \subfloat{ \includegraphics[width=\textwidth, width=1\linewidth]{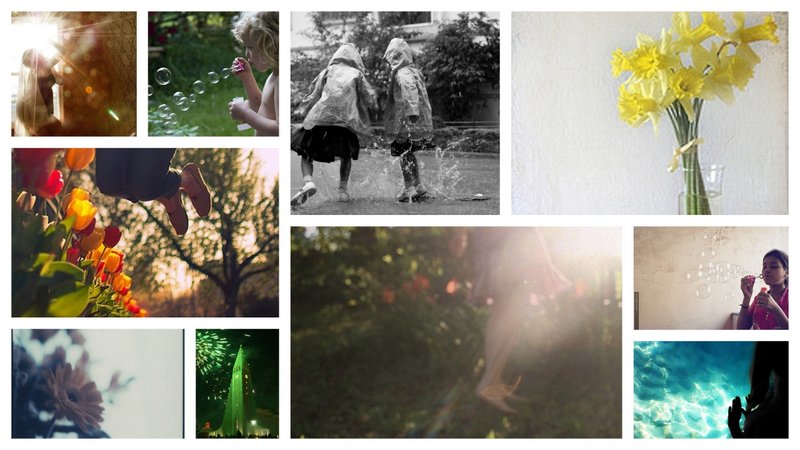}} &
 \vspace{-0.2cm}
 \hspace{-0.5cm}
 \subfloat{ \includegraphics[width=\textwidth, width=1\linewidth]{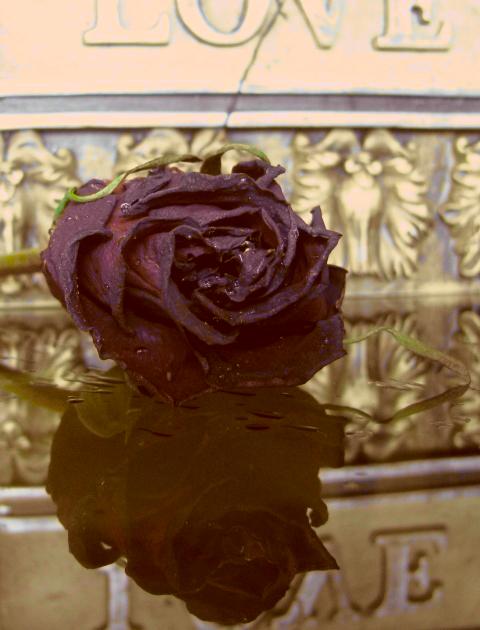}} &
  \vspace{-0.2cm}
  \subfloat{ \includegraphics[width=\textwidth ,width=1\linewidth]{images/sadness_232_1/imgIn.jpg}} &
 \vspace{-0.2cm}
  \hspace{-0.3cm}
 \subfloat{ \includegraphics[width=\textwidth, width=1\linewidth]{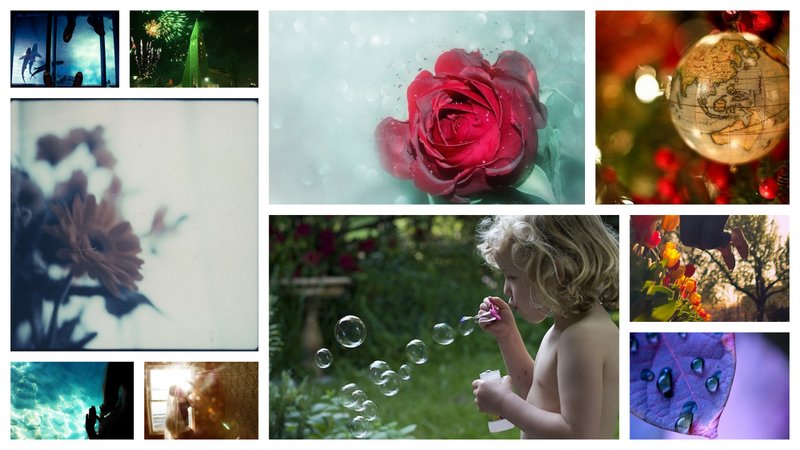}} &
 \vspace{-0.2cm}
  \hspace{-0.5cm}
 \subfloat{ \includegraphics[width=\textwidth, width=1\linewidth]{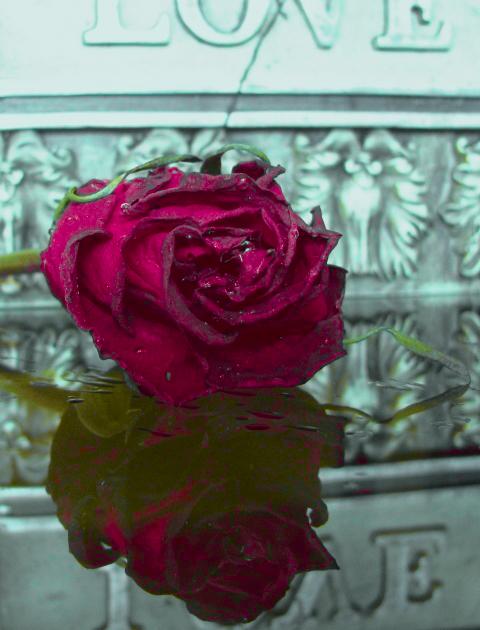}} \\
 \end{tabular}
 
 % -------------------------------------------------------------------------
  \begin{tabular}{m{1.55cm} m{2.35cm} m{1.55cm} : m{1.55cm} m{2.35cm} m{1.55cm}} 
 \vspace{-0.2cm}
  \subfloat{ \includegraphics[width=\textwidth ,width=0.9\linewidth]{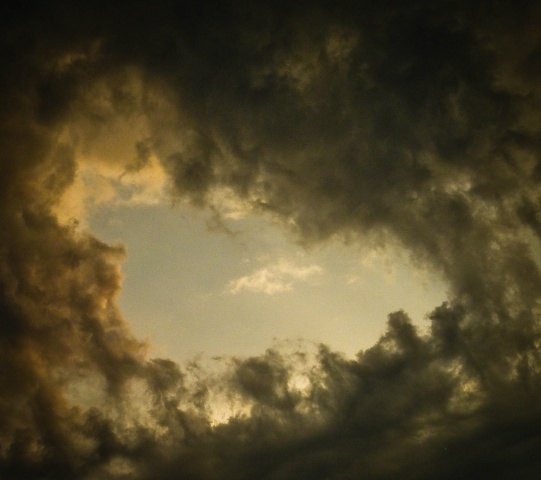}} &
 \vspace{-0.2cm}
  \hspace{-0.3cm}
 \subfloat{ \includegraphics[width=\textwidth, width=1\linewidth]{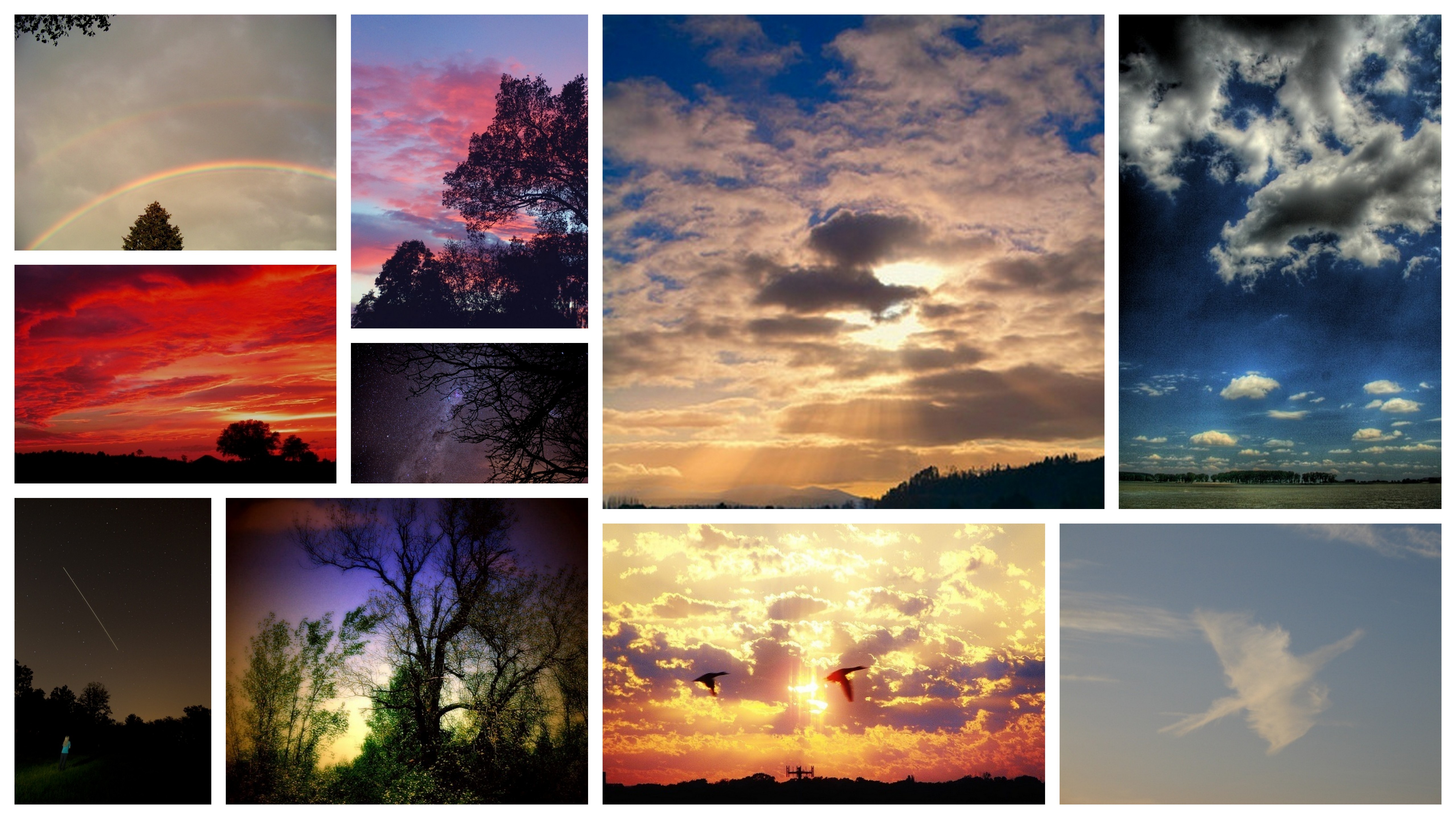}} &
 \vspace{-0.2cm}
  \hspace{-0.5cm}
 \subfloat{ \includegraphics[width=\textwidth, width=0.9\linewidth]{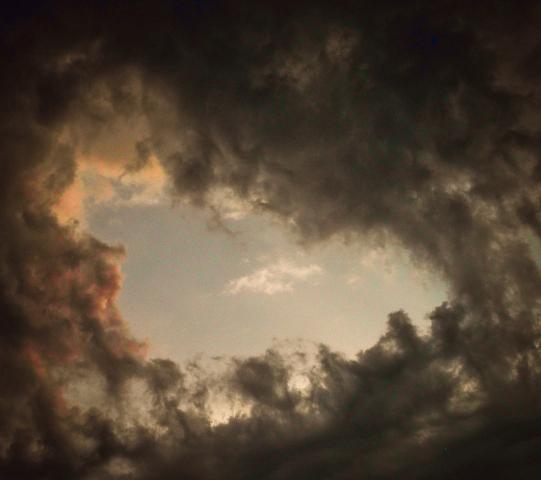}} &
  \vspace{-0.2cm}
  \subfloat{ \includegraphics[width=\textwidth ,width=0.9\linewidth]{images/fear_282_1/imgIn.jpg}} &
 \vspace{-0.2cm}
  \hspace{-0.3cm}
 \subfloat{ \includegraphics[width=\textwidth, width=1\linewidth]{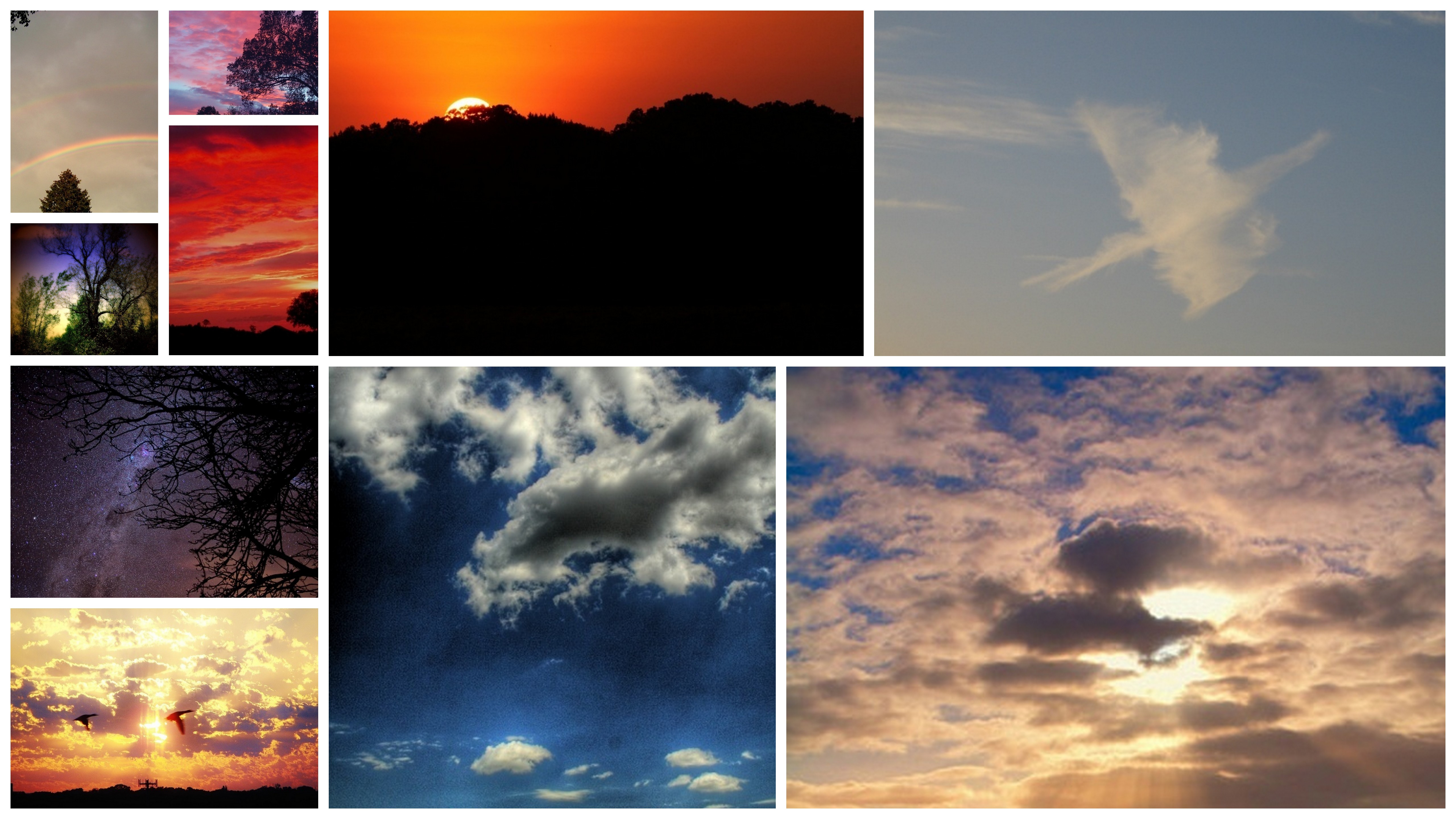}} &
 \vspace{-0.2cm}
  \hspace{-0.5cm}
 \subfloat{ \includegraphics[width=\textwidth, width=0.9\linewidth]{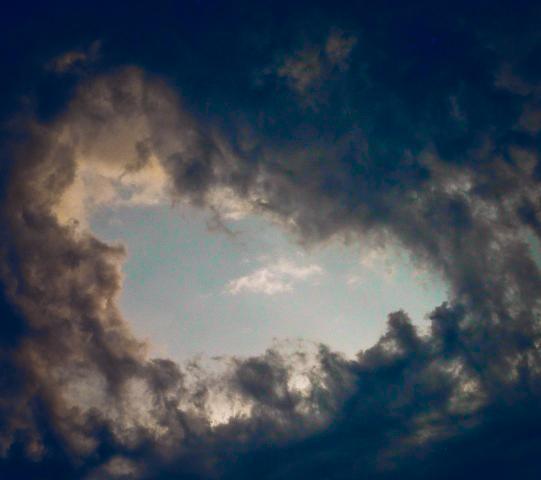}} 
 \end{tabular}
 \vspace{0.2cm}
 % -------------------------------------------------------------------------
\caption{\small{Left-column: transformation results using fc6 features of Alexnet network.Right-Column: shows results with fc7 features. The transformation direction First-Row: Sad to Joy and Secon-Row: Fear to Joy.}\vspace{-0.3cm}}\label{fig:emotionAblation}
\vspace{-0.5cm}
\end{figure*}
\end{center}
% -------------------------------------------------------------------------
\vspace{-1cm}
\subsection{Which layer of CNN to use?}
Our decision to choose fc7 layer of the AlexNet was based on multiple factors, including the size of the of the output from the layer and how much information is captured by that layer. The output layer quantize all the information to the object probability, this would have reduced similarity measure to just counting similar objects in the images. We wanted to measure similarity on the basis of spatial, context and content information. In order to identify the top layer of AlexNet network which has features that are more appropriate for selecting target images and performing emotion transformation, we perform emotion transformation using features from two top layers, named fc6 and fc7, separately. The \mFIG{\ref{fig:emotionAblation}} shows the results of this experiment on two images. These results clearly depict that fc7 features select more appropriate target images and hence generate transformation results that are more close to the target emotion and are more appealing. 

\vspace{-0.4cm}
\section{Failure Cases and Question of High Level Concepts}
\begin{center}
\begin{figure*}[h!]
\center
\hrule
 \begin{tabular}{m{1.55cm} m{2.35cm} m{1.45cm} : m{1.55cm} m{2.35cm} m{1.45cm}} 
\small{\textbf{Input Image}} & \small{\textbf{{Target Distribution}}} & \small{\textbf{{Output Image}}} &
\small{\textbf{Input Image}} & \small{\textbf{{Target Distribution}}} & \small{\textbf{{Output Image}}} \\ [0.5ex] 
 \end{tabular}
  \hrule
 
  % -------------------------------------------------------------------------
 \begin{tabular}{m{1.55cm} m{2.35cm} m{1.55cm} : m{1.55cm} m{2.35cm} m{1.55cm}} 
 \vspace{-0.2cm}
  \subfloat{ \includegraphics[width=\textwidth ,width=1\linewidth]{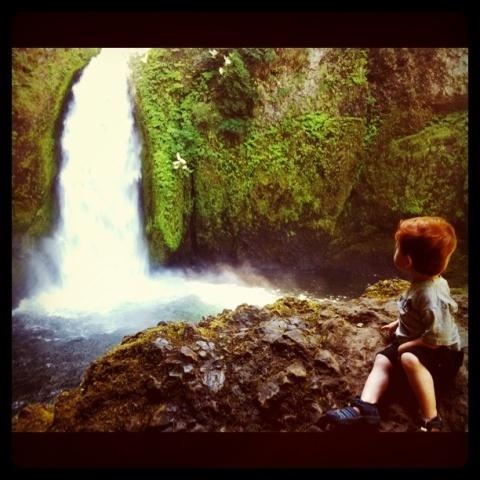}} &
 \vspace{-0.2cm}
 \hspace{-0.3cm}
 \subfloat{ \includegraphics[width=\textwidth, width=1\linewidth]{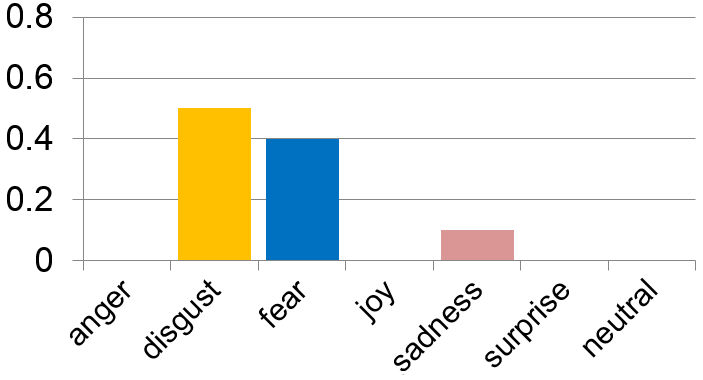}} &
 \vspace{-0.2cm}
 \hspace{-0.5cm}
 \subfloat{ \includegraphics[width=\textwidth, width=1\linewidth]{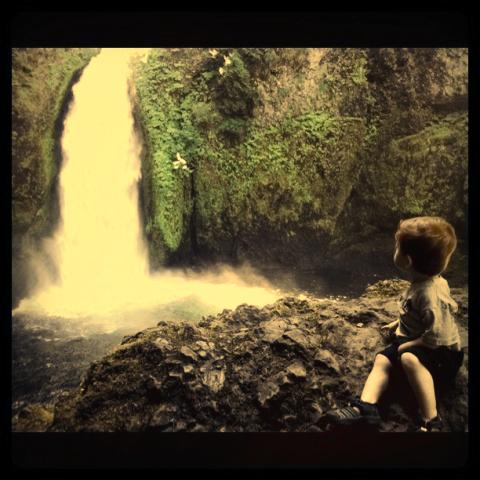}} &
  \vspace{-0.2cm}
  \subfloat{ \includegraphics[width=\textwidth ,width=1\linewidth]{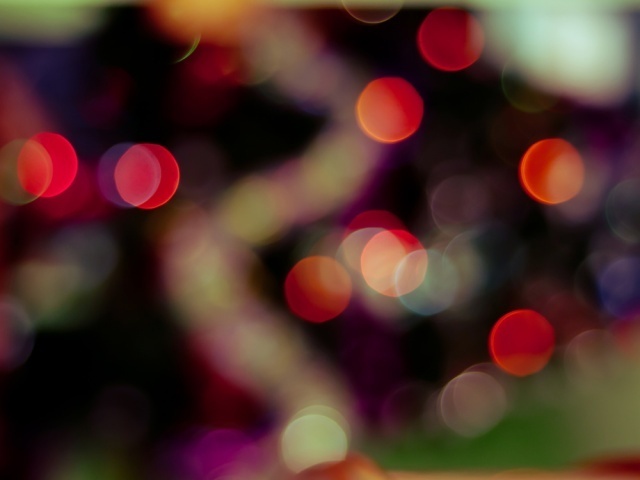}} &
 \vspace{-0.2cm}
  \hspace{-0.3cm}
 \subfloat{ \includegraphics[width=\textwidth, width=1\linewidth]{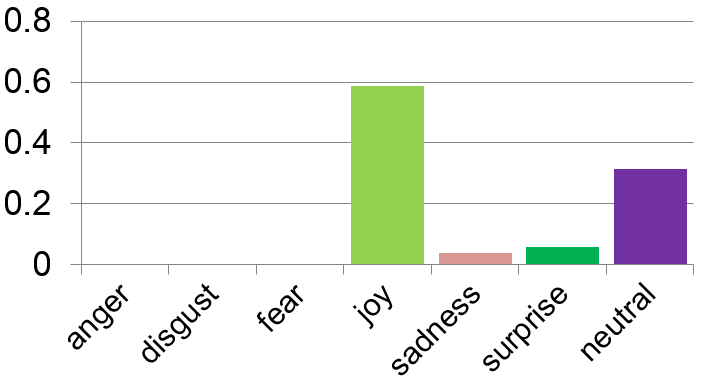}} &
 \vspace{-0.2cm}
  \hspace{-0.5cm}
 \subfloat{ \includegraphics[width=\textwidth, width=1\linewidth]{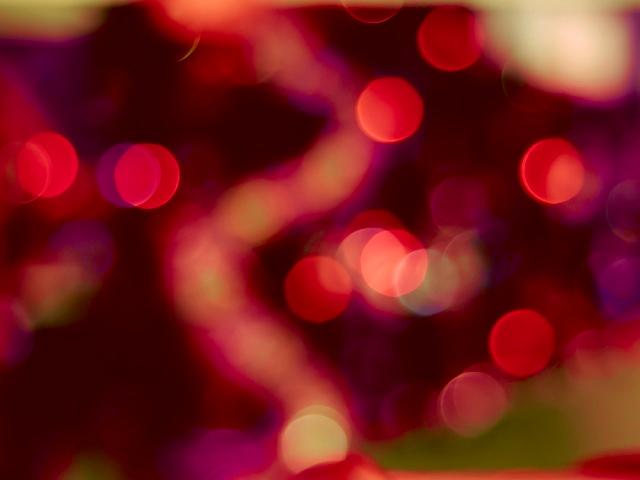}} \\
  \vspace{-0.2cm} 
  \small{ Joy} &  & \small{ Disgust} & \small{ Neutral} &  & \small{ Joy} \\
 
 \end{tabular}
 
 % -------------------------------------------------------------------------

  \begin{tabular}{m{1.55cm} m{2.35cm} m{1.55cm} : m{1.55cm} m{2.35cm} m{1.55cm}} 
 \vspace{-0.2cm}
  \subfloat{ \includegraphics[width=\textwidth ,width=0.9\linewidth]{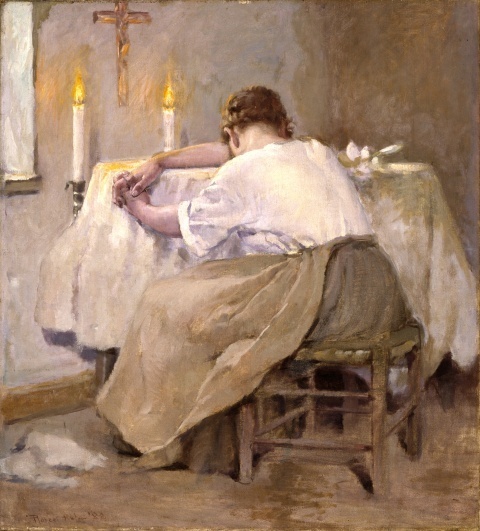}} &
 \vspace{-0.2cm}
  \hspace{-0.3cm}
 \subfloat{ \includegraphics[width=\textwidth, width=1\linewidth]{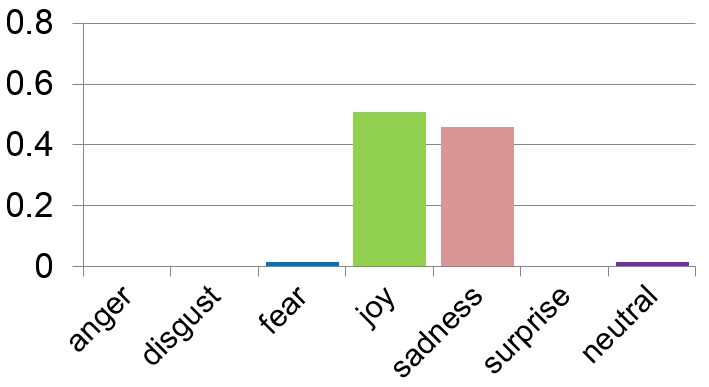}} &
 \vspace{-0.2cm}
  \hspace{-0.5cm}
 \subfloat{ \includegraphics[width=\textwidth, width=0.9\linewidth]{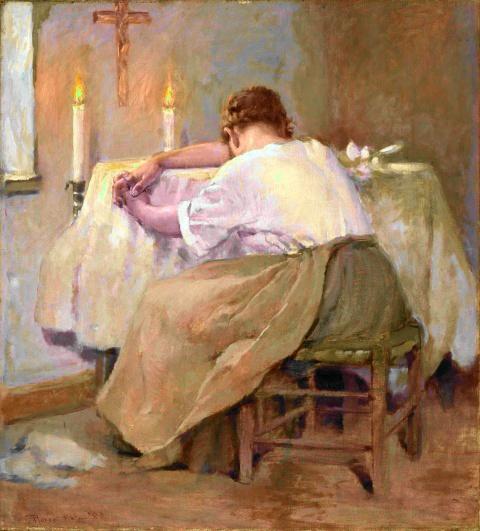}} &
  \vspace{-0.2cm}
  \subfloat{ \includegraphics[width=\textwidth ,width=0.9\linewidth]{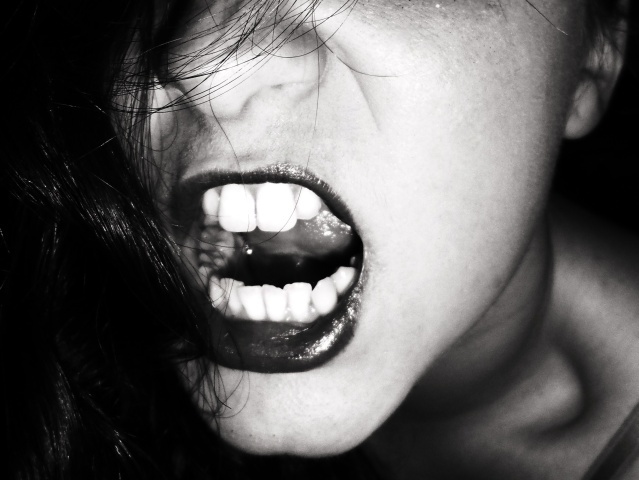}} &
 \vspace{-0.2cm}
  \hspace{-0.3cm}
 \subfloat{ \includegraphics[width=\textwidth, width=1\linewidth]{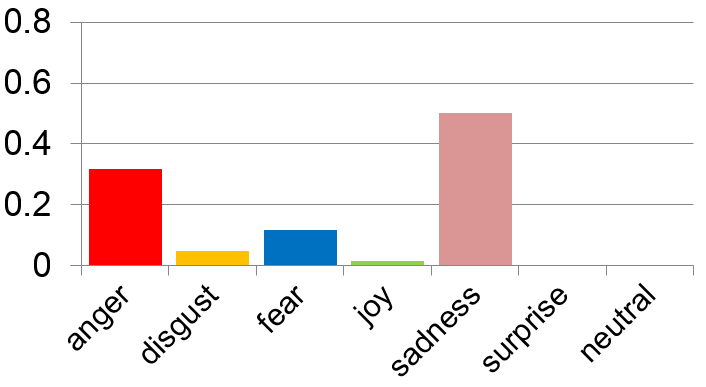}} &
 \vspace{-0.2cm}
  \hspace{-0.5cm}
 \subfloat{ \includegraphics[width=\textwidth, width=0.9\linewidth]{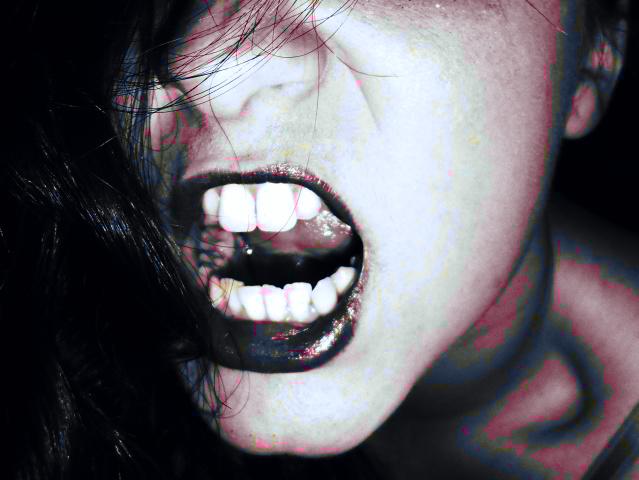}} 
   \vspace{-0.2cm} \\
  \small{ Sadness} &  & \small{ Joy} & \small{  Anger} &  & \small{  Sadness} \\
 \end{tabular}
 \vspace{-0.5cm}
 % -------------------------------------------------------------------------
 \caption{\small{Failure examples of Emotion transformation}\vspace{-1cm}}\label{fig:emotionTransFail}
\end{figure*}
\end{center}

As visible form the results in \mFIG{\ref{fig:emotionTransFail}}, one cannot transfer any image to incite all the emotions. One of the reason is as explained by Ali et al. in \cite{affectHlc2017}, the correlation between the high level concepts in the images and their corresponding elicited emotions, which means that content of the image defines and restricts the spectrum of the emotions that could be elicited from the image. Since we are only transforming low level features (color features) we cannot escape that spectrum. For example, no amount of automatic color and texture manipulation can transform the image of a sad girl to one arousing joyous feelings, without HLC dependent manipulations. Similarly, the images containing neutral concepts (objects having no direct emotion label attach to them) can not be transformed to any other emotion as represented in the second image of first row of \mFIG{\ref{fig:emotionTransFail}}.

\section{Conclusion}
We are witnessing exponential increase in both creation of multimedia, especially photographs, and its sharing on the social network. Digital libraries with millions of images are available for people to share, comment and search on. However, affective analysis of the images, affect base image retrieval or image manipulation to induce emotion has gained much less traction in vision community. 
%However, there are o far, we have seen very little effort in understanding affect of the images, affect based image retrieval or image manipulation to induce the emotion. 
 In this paper, we present a method that allows a user to manipulate an image just by providing the desired emotion distribution. 
 We perform search in the existing database by minimizing over distance between input distribution and emotion distribution of image in database, and content features of the respective images. Histogram for color transformation is constructed from these images selected from database and \cite{pouli2010progressive} is used to transform input image. 
 Since, we use features captured form top layers of CNN trained on object detection and scene identification, we are able to find the images which are similar in content and spatial structure as input image. 
 This allows us to avoid semantic segmentation or patch matching which like previous methods could have restricted us to only few types of images.
 Use of emotion distribution makes the manipulation much more interpretable. We performed a detailed user-study that showed that transformed images generated through our method are better representative of target emotion than the original input image. The failure case and their reasoning is provided, highlighting the limitation of  color-transformation based emotion transfer methods.
\vspace{-0.4cm}
\section{Acknowledgements}
We thank Mr. Junaid Sarfraz for his assistance with the design and implementation of online web portal. The portal was used for conducting user study on perceptual analysis of transformed images.

\bibliography{egbib}
\end{document}